\documentclass[journal]{IEEEtran}
\usepackage{amsmath,amssymb,amsfonts}
\usepackage{algorithmic}
\usepackage{graphicx}
\usepackage{subcaption}
\usepackage{caption}
\usepackage{textcomp}
\usepackage{xcolor}
 \usepackage{url}
\usepackage{comment}
\def\BibTeX{{\rm B\kern-.05em{\sc i\kern-.025em b}\kern-.08em
    T\kern-.1667em\lower.7ex\hbox{E}\kern-.125emX}}

\newcommand{\rev}[1]{{\color{black}#1}} 
\newcommand{\revmin}[1]{{\color{black}#1}}

\newcommand{\name}{OROS}

\begin{document}

\title{Energy-aware Joint Orchestration of 5G and Robots:\\ Experimental Testbed and Field Validation}

\author{\IEEEauthorblockN{
Milan Groshev\IEEEauthorrefmark{1},
Lanfranco~Zanzi\IEEEauthorrefmark{2},
Carmen Delgado\IEEEauthorrefmark{3},
Xi Li\IEEEauthorrefmark{2},
Antonio de la Oliva\IEEEauthorrefmark{1},
Xavier~Costa-P\'erez\IEEEauthorrefmark{2}\IEEEauthorrefmark{3}\IEEEauthorrefmark{4}}
\IEEEauthorblockA{
\IEEEauthorrefmark{1} University Carlos III of Madrid, Madrid, Spain,
Email:\{name.surname\}@uc3m.es\\
\IEEEauthorrefmark{2} NEC Laboratories Europe, Heidelberg, Germany,
Email:\{name.surname\}@neclab.eu,\\
\IEEEauthorrefmark{3} i2CAT Foundation, Barcelona, Spain.
Email:\{name.surname\}@i2cat.net}
\IEEEauthorrefmark{4} ICREA, Barcelona, Spain.
}

\maketitle
\begin{abstract} 
5G mobile networks introduce a new dimension for connecting and operating mobile robots in outdoor environments, leveraging cloud-native and offloading features of 5G networks to enable fully flexible and collaborative cloud robot operations. However, the limited battery life of robots remains a significant obstacle to their effective adoption in real-world exploration scenarios.
This paper explores, via field experiments, the potential energy-saving gains of \emph{\name{}}, a joint orchestration of 5G and Robot Operating System (ROS) that coordinates multiple 5G-connected robots both in terms of navigation and sensing, as well as optimizes their cloud-native service resource utilization while minimizing total resource and energy consumption on the robots based on real-time feedback. We designed, implemented and evaluated our proposed \emph{\name{}} in an experimental testbed composed of commercial \emph{off-the-shelf} robots and a local 5G infrastructure deployed on a campus.
The experimental results demonstrated that \emph{\name{}} significantly outperforms state-of-the-art approaches in terms of energy savings by offloading demanding computational tasks to the 5G edge infrastructure and dynamic energy management of on-board sensors (e.g., switching them off when they are not needed). This strategy achieves approximately $\sim15\%$ energy savings on the robots, thereby extending battery life, which in turn allows for longer operating times and better resource utilization.
\end{abstract}


\section{Introduction}  

The success of outdoor Search and Rescue (SAR) missions depends on the timely coordination of rescue teams on-site to overcome difficulties in locating \revmin{missing} persons and ensuring a prompt \revmin{response}. To minimize rescue teams' exposure to hazardous outdoor environments and enhance search and rescue efficiency, there is an increasing demand for deploying coordinated autonomous robots in mission-critical operations~\cite{rs15133266}. These robots are equipped with heterogeneous sensors and communication capabilities to explore the environment, determine their location, and search for targets or individuals while \revmin{providing} operators with \revmin{real-time} monitoring information. \revmin{Additionally,} Artificial Intelligence (AI) can aid these autonomous systems by enabling more accurate and adaptive decisions based on real-time multi-sensor data streams~\cite{drones7050322}.

In this context, the \revmin{growing} demand for autonomous and coordinated robots is driving a technological shift from standalone robot deployments towards connected autonomous robot platforms. The ubiquitous connectivity, high bandwidth and low latency access \revmin{offered} by the 5\emph{th} generation of mobile networks~(5G), \revmin{combined} with the \revmin{ability} to \revmin{leverage} edge and cloud computing for processing and analytics, enable unprecedented flexibility in deploying modern robotic applications for outdoor scenarios~\cite{5GvsLTE}. Importantly, SAR missions often rely on private 5G networks tailored to the specific requirements of the mission~\cite{5g_private},~\cite{5g_private_1}. This approach addresses the unique challenges of SAR sites, which are frequently located in remote or rural areas—such as mountainous regions—where public 5G coverage is \revmin{typically} unavailable. Private 5G networks ensure that all network resources are exclusively dedicated to the mission, minimizing the risk of resource contention or external fluctuations \revmin{while} providing reliable communication for mission-critical robotic operations.

However, a trade-off arises when pursuing higher \revmin{degrees of} robot autonomy and coordination. While mobility \revmin{remains} the \revmin{primary} source of energy consumption~\cite{Swanborn2020}, \revmin{the} computing and processing of multi-sensor data \revmin{required for} outdoor exploration use cases further increase the energy \revmin{demands} of robotic platforms~\cite{HIRO_NET}. 
Robot operations are heavily constrained by their hardware's computational capabilities and \revmin{associated} energetic aspects. \revmin{Currently}, most of the real-world learning problems faced by robots can only be \revmin{addressed} by \revmin{offloading} dense data to more \revmin{powerful}  computing infrastructure or dedicated edge/cloud computing units~\cite{Mission_Critical_Edge}~\cite{E2E_Reliability}. Therefore, \revmin{ensuring} robust, high-speed, and low-latency communication \revmin{on demand is crucial}.
The current state-of-the-art is evolving by focusing on innovative solutions in individual domains. For example, advancements in the network domain have led to high-speed, low-latency communication, but result in suboptimal robotic performances due to insufficient integration and real-time awareness. Conversely, developments in the robotic domain enhance autonomy and functionality, but often neglect the network aspect of the end-to-end system, limiting real-time communication and coordination capabilities in advanced use cases.
 
Motivated by the need for joint orchestration of the robotics and network domain, our previous work~\cite{OROS_WowMom} \revmin{introduced} \emph{\name{}}, an orchestration framework that uses the knowledge and requirements obtained from the robots (\emph{\textbf{robot-as-a-sensor}}) to seamlessly optimize the network resources and robotic operations. While \emph{\name{}} was extensively evaluated through simulations~\cite{OROS_TNSM}, \revmin{it has yet to be validated} in real-world scenarios. \revmin{Such field validation }is essential to fully characterize its potential, implementation \revmin{challenges} and limitations. \revmin{To address this gap}, this paper \revmin{focuses} on the experimental evaluation of \emph{\name{}} in an outdoor testbed with real mobile robots and local 5G connectivity. Our implementation provides a comprehensive testing environment for experimenting with various network, computation, and robotic configurations. 
Using this testbed, we deploy open-source mobile robots, an object detection robotic service, and a prototype of \emph{\name{}} to optimize robot trajectories and minimize energy consumption.
The main contributions that distinguish this work from the state\revmin{-}of\revmin{-}the\revmin{-}art are:
\begin{itemize}
    \item We perform a comprehensive energy consumption profiling of mobile robot applications.
    \item We develop a cloud-based, containerized cross-field testbed that serves as a playground for testing and validating innovative solutions integrating both network and robotics domains. This testbed also incorporates 5G technology and features dedicated APIs specifically developed for seamless interaction between the network and robotic components.
    \item We build a digital model relying on \revmin{the} ROS-based Gazebo \revmin{s}imulator to emulate the behavior of multiple mobile robots into a digital representation of the deployment environment to facilitate testing and fine-tuning
    of \emph{\name{} before its deployment in the physical environments.}
    \item We design and implement \emph{\name{}} in a real outdoor testbed, integrating both 5G and robot domains, evaluating its performances through a field trial and demonstrating its feasibility in realistic environments.
\end{itemize}

The software and APIs of \emph{\name{}}, the digital model of the test environment, as well as the measurement data from the testbed will be available to the public after acceptance of the paper. This can provide to both the networking and robotic research community a playground for testing and evaluating the proposed solution with real field experiments, in turn fostering openness in this research field and encouraging collaboration and knowledge-sharing.

The remainder of this paper is structured as follows.
Sec.~\ref{sec:framework} presents the needed background on 5G and robot orchestration as well as a short description of \emph{\name{}} and its interaction with the different modules.
Sec.~\ref{sec:POC} introduces the real-world Proof-of-Concept (P\revmin{o}C) testbed and details its main components.
Sec.~\ref{sec:profiling} provides an exhaustive energy profiling based on measurements from two 5G-enabled mobile robots.
Sec.~\ref{sec:outdoor} validates the design principles of our solution and provides a field test validation, highlighting the main benefits derived from our approach.
\rev{Sec.~\ref{sec:scalability} discusses the algorithm complexity, providing insights on improving the overall 
system scalability.}
Sec.~\ref{sec:related} summarizes related works in the field.
Finally, Sec.~\ref{sec:conclusion} concludes this paper.
\begin{figure}[t!]
      \centering
      \includegraphics[trim = 2cm 1.9cm 2cm 0cm , clip, width=\columnwidth ]{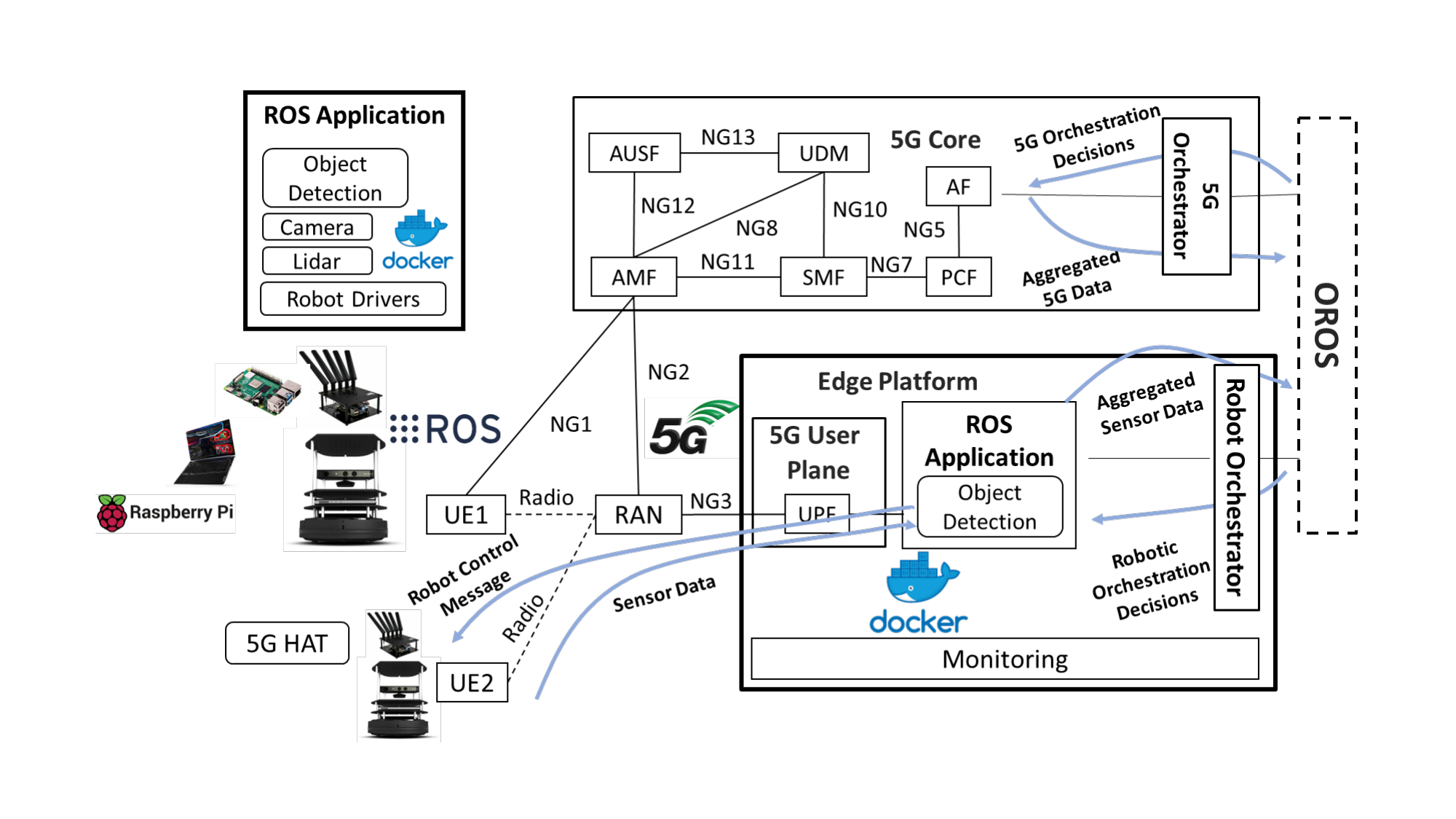}
      \caption{\small Overview of the architectural building blocks.}
      \label{fig:scenario}
      \vspace{-3mm}
\end{figure}
\section{\emph{\name{}} for collaborative robot orchestration}
\label{sec:framework}
We consider a set of ground robots deployed in an unknown outdoor environment, as depicted in Fig.~\ref{fig:scenario}. Wireless communication is provided by a 5G network, where a 5G Radio Access Network (RAN), \revmin{consisting} of \revmin{multiple} base stations (gNBs), \revmin{ensures} radio coverage over the area of interest. The 5G \revmin{infrastructure} may also \revmin{incorporate} an edge and central cloud platform to host robotic applications and 5G network functionalities. \revmin{These components run as virtualized or container-based instances on a shared computing infrastructure.}
To \revmin{achieve} low latency communication with the remote-controlled robots, the robot controller is usually \revmin{hosted at the edge, with a User Plane Function (UPF) deployed at the same location}. \revmin{This allows seamless routing of 5G data plane traffic between the robots and the edge.}

\rev{Although we assume a 5G private network, it is important to mention that \revmin{robotic operations} remains robust even under varying network conditions. \revmin{This ensures that SAR missions can still function if public 5G networks are used, or if private networks experience fluctuations due to wireless propagation challenges~\cite{Singhal2017}.} This robustness is primarily ensured by two key considerations. First, centralized coordinated control for the SAR service operates at lower frequencies (10 Hz or be\revmin{l}ow), making the control process less time-sensitive. Since 5G networks typically provide an average latency of 10-20 ms, such fluctuations remain within tolerable limits. Second, robotic systems are designed with control-theory mechanisms and equipped with Kalman filters to autonomously handle delayed or lost control or sensing packets. These mechanisms ensure operational continuity even in the rare event of network-induced disruptions, providing a robust and resilient solution for mission-critical operations.} 

\rev{Even though robotic systems are highly robust, the dynamic nature of 5G networks and the available resources allocated to robotic applications can fluctuate over time in real-world scenarios~\cite{Singhal2017}. These fluctuations can impact the data rate and latency of robotic applications, affecting their overall performance. For instance, a robot may struggle to transmit data during signal blockage or under poor radio signal conditions. This challenge must be taken into account and addressed by the orchestration framework.} In the context of our work, we identify two different orchestration domains, namely the 5G domain and the robotic domain. In the 5G domain, the orchestration is in charge of the allocation of 5G infrastructure resources to meet individual application and communication requirements. In the robot domain, orchestration is needed to control the multitude of operational modules and sensors installed on the on-board robot platform, to optimize their energy consumption and task performance.
However, so far, great effort has been devoted to independently orchestrating individual domain platforms, e.g. OSM~\cite{OSM}, ONAP~\cite{ONAP} to manage 5G infrastructure resources and ROS~\cite{ROS} to control robot applications and operations. Due to a lack of dedicated data models and interfaces to connect them, there is no real-time interaction between the two domains and hence the two orchestration tasks work independently, with little to no awareness of each other during the operational phases. This often leads to non-optimized decisions. On the one hand, robot operations are based on the wireless communication network and the backhaul infrastructure for computation offloading. Especially enabled by softwarization and Network Function Virtualization (NFV) technologies, modern robot devices generate and therefore require processing large volumes of sensing data by dedicated software applications running on a shared computing infrastructure at the network edge, which may constrain the performances of data processing and task operation. Without proper knowledge of the network performance and resource availability, the robots simply make "blind" decisions on their tasks assuming perfect network connectivity, which may fail or degrade robot operations in bad radio link conditions. \rev{For instance, without sufficient network resources and stable link conditions, the robot is not able to transmit high bandwidth data or offload high processing tasks to the edge. } On the other hand, mobile networks may require real-time information on the amount of generated robotic data and traffic volume, the robot states (e.g., battery level, onboard sensor states, etc.) and their tasks, so as to be properly configured to allocate sufficient resources to ensure the bandwidth and latency requirements of various robotic applications. 

Without joint orchestration, these software instances would always remain active, impacting the overall energy consumption of the robots. To overcome this problem, we advocate for the adoption of joint 5G and robot orchestration solutions to guide the overall life-cycle management of software instances, provisioning of dedicated cloud computing resources, and instruct context-aware robot motion planning, pursuing energy savings strategies.

\subsection{Robot Orchestration}
In this paper, we build on the ROS framework and specifications~\cite{ROS} for the control and orchestration of robots and the onboard robotic applications. ROS is an open-source robotics middleware that provides a meta-operating environment for developing and testing multi-vendor robotics software. In ROS, each software component is called ROS node. Moreover, ROS provides a publish-subscribe messaging framework via a specific node, namely ROS master. By connecting to the ROS master, \revmin{the} ROS nodes can register and locate each other. Once registered, nodes can exchange data via configurable topics peer-to-peer. 

The \emph{Robot Orchestrator} is \revmin{responsible for managing and coordinating} multi-robot systems \revmin{within the ROS framework ensuring seamless integration and control of various ROS applications.}
It is structured into three layers: the application layer, ROS client layer, and ROS middleware layer. The application layer hosts a variety of robotic applications with run-time application programming capabilities. The ROS client layer exposes a set of ROS client APIs~\cite{ROS_RCL_APIs}, derived from the built-in ROS client libraries, enabling multi-language support (e.g., C, C++, Python) for robotic application development. The ROS middleware layer offers a set of APIs~\cite{ROS_RWP_APIs} to enable compatibility with different low-level communication protocols, \revmin{facilitating} distributed data and service exchange. Through these API interfaces, the Robot Orchestrator translates high-level application logic into executable instructions, which are then dispatched as ROS command messages via its Southbound Interface (SBI) to control and coordinate multiple robots effectively.

\subsection{5G Orchestration}
In its fundamental role, a 5G Orchestrator is responsible of the allocation and management of the 5G infrastructure resources, encompassing both network resources, which facilitate robot communication and application data transmission, and computing resources, which host and execute robotic control plane applications.
Beyond resource management, the 5G \revmin{O}rchestrator determines the optimal placement strategy of the robotic applications. This strategy allows for direct deployment on robot devices, offloading to \revmin{the} edge or cloud infrastructure a distributed execution model, depending on the capabilities of the underlying software instance.

Effective application placement requires proactive resource allocation to ensure optimal provisioning of computing, memory, and storage resources across robots, edge nodes, and cloud platforms. Moreover, the 5G Orchestrator is responsible for the lifecycle management of robotic applications, encompassing onboarding, instantiation, monitoring, and enforcement operations (e.g., automatic scaling and self-healing mechanisms). These functionalities enable dynamic resource adaptation in response to network fluctuations and robot mobility, ensuring seamless orchestration even under varying operational conditions.

The 5G \revmin{O}rchestrator can be built relying on existing open source orchestrator platforms such as Open Source MANO (OSM), or leveraging on research-based open source orchestration platforms developed for instance in \cite{5GT_arch} \cite{5Growth-commag} that runs automatic and is tailored for different vertical applications. 

\subsection{\name{}}
\label{subsec:\name{}}
To fill the gaps between these two domains, we proposed \emph{\name{}} \cite{OROS_WowMom}, a joint orchestration solution for the robotic and 5G ecosystem, to control ROS-driven collaborative connected robots in 5G networks. \emph{\name{}} uses the \emph{\textbf{robot-as-a-sensor}} concept where the knowledge and requirements from the robots deployed in the outdoor environment are used together with the network's real-time status to effectively link and coordinate the operations of both the robot and 5G networks. 

\begin{figure}[t!]
       \centering
       \includegraphics[width=0.99\columnwidth]{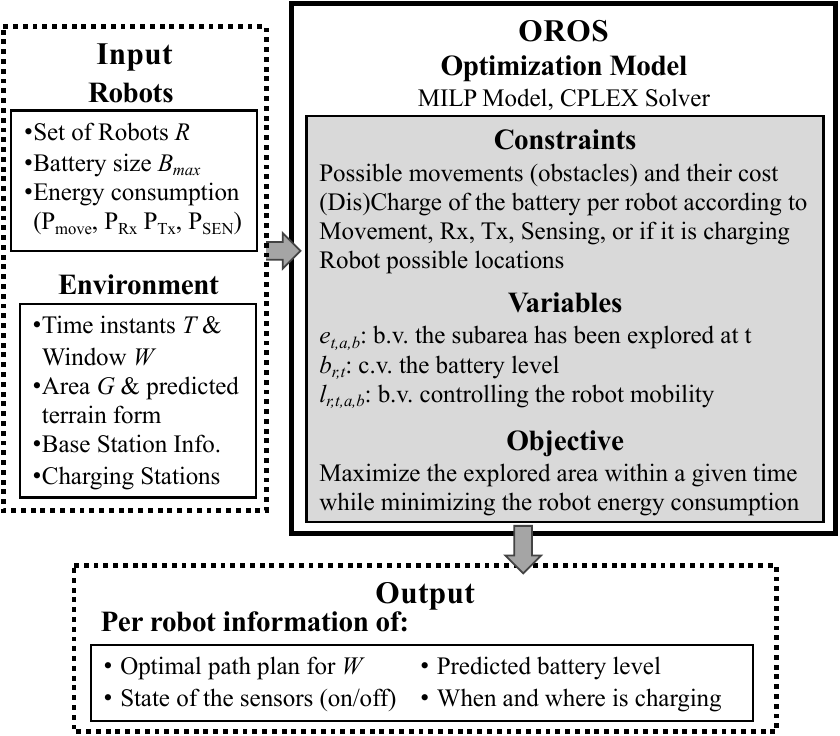}
       \caption{\small \emph{\name{}} Optimization model structure.}       
       \label{fig:optimization}
\end{figure}

The orchestration module acts as a coordination entity between robotics and 5G domains, generating \textit{Robotic Policies} and \textit{5G Policies} based on real-time data from both the Robot and 5G Orchestrators. This includes robots' operational metrics and network resource availability (i.e., robots' speed, their current location, instantaneous energy usage and battery levels, available radio and computing resources). The Robot Orchestrator translates these into executable commands, adjusting robot activities to optimize resource usage. Meanwhile, in the 5G sphere, the Intent Engine adjusts network configurations in response to these policies, reallocating resources like RAN and core networks and managing robotic application migration, in line with ETSI IFA 005 standards \cite{ifa005}.
%%%%%%%%%%%%%
In order to orchestrate the robot and 5G-domain decisions, \emph{\name{}} iteratively solves a Mixed-Integer Linear optimization problem maximizing energy efficiency and robot navigation, as depicted in Fig.~\ref{fig:optimization} which specifies the model structure and both inputs and outputs of the problem.
We assume the system evolves according to a discrete set of time instants denoted by \textbf{$\mathcal{T} = \{t_1, \dots, t_{|\mathcal{T}|} \}$}.
The environment is modeled as a grid $\mathcal{G} =\{g_{a,b}, \forall (a,b) \}$, with robots $\mathcal{R}=\{r_1, \dots, r_{|\mathcal{R}|} \}$ needing to explore each grid element potentially until a target is located. The model calculates robots path planning and the corresponding energy consumption, aiming to minimize energy consumption while maximizing the exploration rate.
As detailed in the Input block of Fig.~\ref{fig:optimization}, the key inputs of the algorithm include the set of robots \textbf{$\mathcal{R}=\{r_1, \dots, r_{|\mathcal{R}|} \}$}, each with a battery limit $B_{max}$ and the set of available sensors, e.g., camera, lidar, etc., as well as the estimated energy consumption per component. To address this, we introduce the variable $P_{move}$ which considers the obstacles and terrain form to account for the energy needed to navigate, the transmitting and receiving power $P_{TX,a,b}$ and $P_{RX}$ to consider the energy consumed by the robot to send/receive traffic through the radio interface, and $P_{SEN}$ which represents the energy consumed by the robot sensors and the corresponding local data processing, as well as the energy consumption coming from the computing infrastructure locally running on the robot. 
\rev{An important feature to be considered when orchestrating multiple robots is the heterogeneity of robots' capabilities in terms of sensors, battery, and related energy consumptions, as also investigated by prior studies demonstrating the impact of such heterogeneity on key metrics such as power consumption~\cite{Romero2024}. Our \textit{OROS} framework accounts for such variability providing the flexibility to accommodate different robot and sensor configurations. However, to streamline the notation and reduce the clutter, we have omitted the robot-specific suffix~$r$, which can be reinstated as needed to reflect specific deployments.}
Moreover, the optimization problem requires several environmental-related information in input, including
\textbf{$\mathcal{G} =\{g_{a,b}, \forall (a,b) \}$} representing the area of interest, \textbf{$(g_{a_{BS},b_{BS}} ) \in \mathcal{G}$} representing the serving base station location, and the information of the charging stations (i.e., location and charging rate).
As detailed in the \emph{\name{}} Optimization Model block of Fig.~\ref{fig:optimization} we introduce $e_{t,a,b}$ as a binary variable indicating if the 2D area unit $g_{a,b}$ has been already explored at time $t \in \mathcal{T}$, or not.
Finally, let $l_{r,t,a,b}$ be a binary decision variable to control the robot's mobility. Its value gets positive if robot $r$ is at position $g_{a,b}$ at time instant~$t$.
 \begin{figure}[t!]
       \centering
        \includegraphics[width=.80\columnwidth]{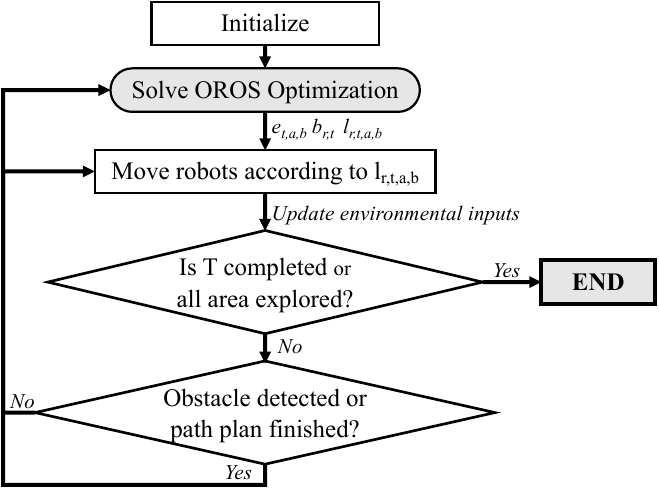}
       \caption{\small \emph{\name{}} algorithm.}
       \label{fig:algorithm}
\end{figure}
To increase the chances of detecting the target object (or person) in SAR operations, the goal is to maximize the explored area within a given time frame while minimizing the energy consumption of the robot platform. Therefore, as detailed in \cite{OROS_TNSM}, the problem constraints can be summarized as:
\begin{itemize}
    \item The robot's movement at any given time instant must be restricted by the positions of \revmin{the} detected obstacles, only allowing for adjacent positions.
    \item The battery discharge rate for each robot must be calculated based on the movement and activity performed, to ensure it stays within the operating limits of the battery. 
    \item Whenever the robot is charging at the Charging Station, its battery increases at the charging rate.
    \item The exploration progress among multiple robots must be monitored to ensure \emph{\name{}} makes informed decisions on the possibility of saving energy when a robot explores an already visited area.
    \item The status of sensors, camera, processing units, and transmission elements must be managed based on the area and time instant, ensuring that these components are turned off if those areas have already been explored to save energy and improve the operational limits. 
\end{itemize}

A solution to the problem includes several outputs, as shown in the Output block of Fig.~\ref{fig:optimization}. Firstly, it provides the optimal positions of the robots, detailing their path plan. Secondly, it indicates the state of the sensors, specifying whether they should be turned on or off. Lastly, it predicts the battery level of each robot at every instant (whether it is charging or discharging), which is crucial for detecting potential errors or terrain changes, such as hills. For instance, if the predicted battery level is significantly higher than the reported level, this discrepancy may indicate that the robot encountered a slope.

The problem outlined above can be approached in various ways. An \emph{offline} approach assumes complete knowledge of the input variables and solves a single instance of the problem for the entire \revmin{time frame}, as discussed in~\cite{OROS_WowMom}. Although this method provides a benchmark for \revmin{the} optimal performance, its application in real-world scenarios is limited to specific situations where such comprehensive information is available. Consequently, in this work, we adopt the \emph{online} algorithm approach demonstrated in~\cite{OROS_TNSM}, solving the problem iteratively, as illustrated in Fig.~\ref{fig:algorithm}.
Once the algorithm is initialized, it solves the problem for the next $W$ steps, where $W$ represents the desired time window. 
\rev{Although this solution is not optimal, it reduces the computational complexity of the MILP model, enabling the system to accommodate a larger number of robots and areas. } Based on the output path planning, the robots will move accordingly. The algorithm continues to operate as long as there are areas to explore or time intervals to complete. Additionally, any robot-related events, such as object detection or terrain slopes, or the completion of the path planning for the $W$ steps, trigger a new solver task.
\rev{Determining the optimal value of $W$ is not trivial, as it depends on factors such as area size, the number of robots, and obstacle density, as previously discussed in \cite{OROS_TNSM}. However, it can be estimated prior to deployment.}

 %v1 or v2?
\section{Proof-of-Concept Design and Implementation}
\label{sec:POC}

This section presents a PoC design and implementation of \revmin{the} \textit{OROS} system that enables experimentation with robotic deployments and evaluation of resource allocation and orchestration algorithms. The testbed follows a cloud-native approach, where each building block is implemented as a virtual network or robot function, including:
\begin{itemize}
  \item \textit{Mobile Robot Application}, which implements an object detection service composed of $5$ virtual functions using ROS and \revmin{d}ocker. We also evaluate the energy consumption of each of the robot sensors.
  \item \textit{Robot Orchestrator} implements a set of ROS APIs that \emph{i)} expose the robot application information (e.g., robot odometry, video stream, lidar point cloud, object detection) and \emph{ii)} allow control of the robot application (e.g., teleoperation control, lidar configuration, camera resolution, etc.).
  \item \textit{5G Orchestrator} implements a set of 5G and 
\revmin{d}ocker APIs that \emph{i)} expose the 5G and compute-related information (e.g., 5G wireless channel status, used CPU, RAM) and \emph{ii)} control the network and compute infrastructure (e.g., life-cycle management of the virtual functions, configuration of radio parameters).
  \item \textit{Prototype of \name{\textit{}}} orchestration solution according to Sec.~\ref{subsec:\name{}}, that enables inter-domain interaction and effectively implements joint-optimization strategies.
\end{itemize}

\subsection{Testbed Platform}
%HW setup of the testbed
The setup of our experimental testbed hosts two ROS-compatible Kobuki TurtleBot2 S2 robots\footnote{\label{robot} https://www.turtlebot.com/turtlebot2/}\revmin{,} equipped with a RPLIDAR A3 lidar\footnote{\label{lidar} https://www.slamtec.com/en/Lidar/A3} and an Orbec Astra 3D camera\footnote{\label{camera} https://shop.orbbec3d.com/Astra}\revmin{,} that act as our robot-as-a-sensor to gain knowledge from unknown environments. Additionally, we installed a 5G Hardware Attached on Top (HAT)\footnote{\label{radio_hat} https://www.waveshare.com/sim8200ea-m2-5g-hat.htm} to enable 5G connectivity. Each robot is also equipped with a set of dedicated computing resources provided by a laptop running on Ubuntu operating system (OS). It is worth mentioning that each robot has a different laptop characterized by \revmin{a} different processing power (Intel Core i5-1335U at 3.4 GHz vs. Intel Core i7-1250U at 4.7 GHz) and battery capacity (55 Wh (4-cell) vs. 60 Wh (4-cell)). 
The laptops execute software components of the robotic application as 
\revmin{d}ocker containers. Both 5G HATs are connected to the 5TONIC~\cite{5TONIC} provided local 5G network\revmin{,} which includes an outdoor radio unit (Radio 4408 B78R) from Ericsson connected to the 5G RAN, an edge platform hosting the 5G User Plan Function (UPF) and ROS applications, and a cloud platform that runs the remaining 5G core functions. The local 5G network is deployed in standalone (5G SA) mode. 
The edge platform runs a DELL PowerEdge C6220 server, equipped with 96GB of RAM, 2x Intel Xeon E5-2670 (2.6 GHz) and NVIDIA GeForce RTX 2080 Ti GPU, and a Dell PowerEdge R630 server, equipped with 128GB of RAM and 2x Intel Xeon E5-2620. It offers edge computing resources to offload heavy computational tasks, facilitate robot coordination and, optionally, process the robot-originated data.

\subsection{Robotic Application and Orchestrators}
\begin{figure}[t!]
      \centering
      \includegraphics[clip, trim = 3cm 2cm 1cm 1cm, width=\columnwidth]{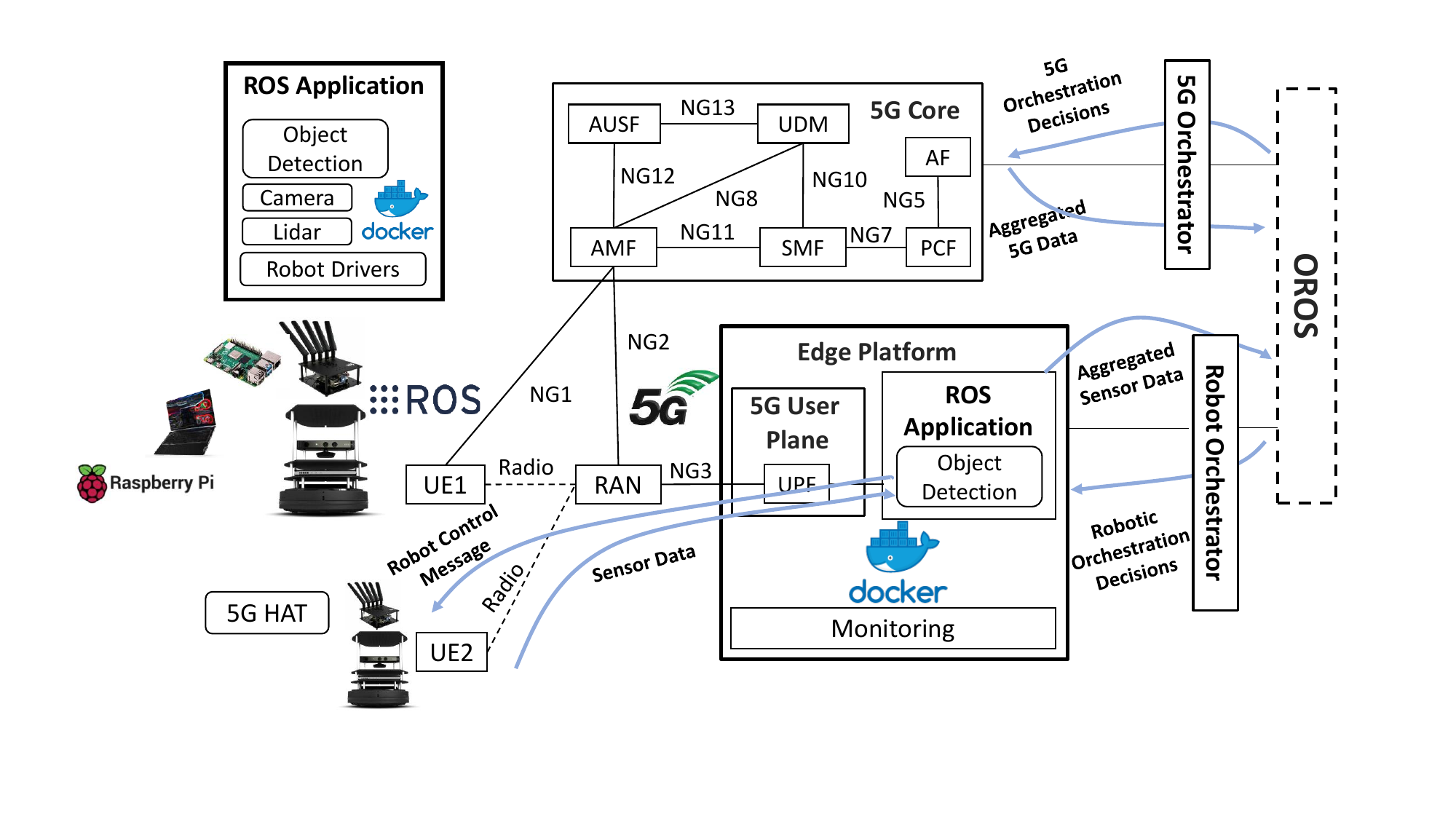}
      \caption{\small \rev{Proof-of-Concept Implementation.}}
      \label{fig:hw_sw_setup}
\end{figure}

%SW setup
\revmin{Fig.~\ref{fig:hw_sw_setup} shows the} robotic applications, each composed of one or more robot virtualization functions, \revmin{that} can be placed/distributed between the robot computing platform and the edge platform. Virtualized robotic functions are implemented \revmin{using} ROS1 and 
\revmin{d}ocker. Basic robot sensors (lidar, camera) and actuators (robot drivers) run as containerized 
\revmin{d}ocker functions in the robot's laptop. Furthermore, the robot software stack is also composed of a virtual computer vision function used for object detection that is implemented using Yolo ROS~\cite{bjelonicYolo2018}. The object detection virtual functions can be deployed locally on the robot or offloaded to the edge platform. Each of the virtualized robotic functions is composed of multiple ROS nodes (software modules) that interact with each other via ROS topics or services. After booting, the virtualized robotic functions contact the ROS master to register and subscribe to the ROS topics/services of interest. 

The Robot and 5G Orchestrator are deployed in separate computing nodes and \revmin{are} responsible for the management of the robot applications and the computing and communication infrastructure, respectively. \rev{For simplicity in this proof-of-concept implementation}, the Robot Orchestrator is developed in Python and deployed using 
\revmin{d}ocker at the edge platform. It uses the ROS API to aggregate and expose relevant \revmin{robot sensor data} such as robot odometry, lidar laser scans or the bounding boxes of the detected objects. In addition, the Robot Orchestrator provides a set of SBI APIs that enables remote control of the robots by translating the high-level movement decisions to low-level ROS navigation commands. 

The 5G Orchestrator is also developed in Python, and it is in charge of managing the resources and actions over the 5G infrastructure. REST-based communication allows \emph{\name{}} to instantiate, deploy, and control the life-cycle of robotic application instances, including starting and stopping individual sensor devices on the robots. This is achieved by employing the 
\revmin{d}ocker Python API. \rev{Similar to the Robot Orchestrator, the current implementation of the 5G Orchestrator is simplistic to facilitate experimental analysis. Future iterations of our testbed will incorporate advanced 5G and Robot \revmin{O}rchestrators to enable more detailed performance evaluations and scalability tests.}

Finally, on the edge platform, we develop an \textit{online} prototype of \emph{\name{}} that leverages the Robot and 5G Orchestrator APIs to jointly coordinate networking resources while providing navigation instructions during the exploration phases. The \emph{\name{}} prototype is developed in C++ programming language and follows the architecture presented in Fig.~\ref{fig:scenario}. It acts as the central decision-making entity to coordinate the actions of the Robot Orchestrator and the 5G Orchestrator.

\subsection{\rev{Real-time data exchange and workflow}}
\rev{In our experimental setup, the real-time data exchange between the robotic and 5G domains is critical for ensuring efficient and adaptive operations in dynamic and unknown environments. The robots deployed in the field continuously collect environmental information using the RPLIDAR A3 lidar and Orbec Astra 3D camera. The \revmin{lidar} sensor, operating at a sampling rate of 25,000 samples per second, generates approximately 1 Mbps of data. The camera sensor, configured for RGB and depth streaming at 720p resolution and 30 FPS, requires between 5 and 15 Mbps when compressed using H.264. Together, these sensors demand a stable uplink bandwidth of approximately 16 Mbps per robot, with low latency ($<$50 ms) and minimal packet loss ($<$1\%) to maintain real-time performance.

The virtualized robotic functions residing in the robot convert the raw sensor data into ROS messages. Using the 5G HAT, the data is transmitted to the edge platform where the Robot Orchestrator resides. The Robot Orchestrator aggregates this data and exposes it to \textit{OROS} via the SBI API for further analysis. This enables \textit{OROS} to maintain a real-time understanding of the robots’ surroundings, which is crucial for coordinated decision-making. Simultaneously, the 5G Orchestrator, via a REST API, exposes information about the status and resource usage of the 5G virtual network functions and robotic application instances. This information allows \textit{OROS} to combine the robot application knowledge with the 5G infrastructure and achieve dynamic control of the robot sensors.
\textit{\name{}} processes the upstream sensor data, along with the robot and 5G network service status, to make adaptive decisions regarding robot operations. These decisions are communicated to: i) the Robot Orchestrator as high-level control commands, such as movement direction, speed, or exploration area, and ii) the 5G Orchestrator as infrastructure control messages to control the life cycle of robotic application instances, including starting and stopping individual sensor devices on the robots. For downstream control, the network must support low-bandwidth commands (10–20 KB/sec per robot) with latency under 50 ms to ensure responsive and accurate robot behavior.} 

\subsection{Robotic Simulator and Data Collection} 
\label{subsec:simulator}
\begin{figure}[t!]
      \centering
      \includegraphics[trim = 0cm 0cm 0cm 0cm, clip, width=\columnwidth ]{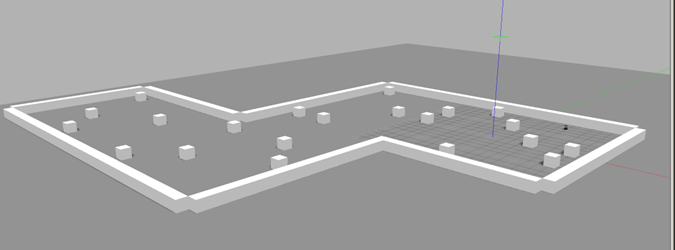}
      \caption{\small Virtual environment with random obstacle positions.}
      \label{fig:gazebo}
      % \vspace{-4mm}
\end{figure}

As mentioned before, our testbed has an application server that runs in the edge platform enabled with an NVIDIA GeForce RTX 2080 Ti GPU and 600 GB of storage. We use this server to deploy the ROS-based Gazebo simulator that facilitates comprehensive testing and evaluation of navigation algorithms, enabling developers to fine-tune and validate their implementations before deployment in physical environments. This iterative process helps to streamline the development phase and optimize the performance of robotic systems, ultimately enhancing their capability to navigate autonomously and efficiently in diverse environments. Therefore, to guide the development phase of \emph{\name{}} and preliminary assess its performance, we make use of the Gazebo software to emulate the behavior of multiple mobile robots into a digital representation of the real deployment environment (described in detail in Sec.~\ref{sec:outdoor}). This is shown in Fig.~\ref{fig:gazebo}. This digital model serves as a faithful emulation of the real-world deployment settings and allows testing heterogeneous obstacle placements and the system APIs to control robots and acquire real-time feedback. To favor reproducibility, we make this digital environment accessible to other researchers and developers to replicate our experiments, fostering transparency and encouraging collaboration and knowledge-sharing within the robotics and wireless communities\footnote{Online Available: \emph{Data will be made public upon acceptance of the manuscript.}}.
To gather monitoring metrics from the robot application, computing and communication infrastructure, we use an InfluxDB time-series database instance deployed in the application server to store the monitoring metrics data. The Robot and 5G Orchestrator periodically gathers and stores data in the database. To ensure robot function synchronization and ease the processing of multi-host data sources, we ensure the time synchronization of all hosts using the Precision Time Protocol (PTP).  

\section{Robot Energy Consumption Profiling}
\label{sec:profiling}

\begin{figure}
    \centering
    % \vspace{3 mm}
    \includegraphics[clip, trim = 4cm 7cm 6.5cm 2.5cm, width=\columnwidth]{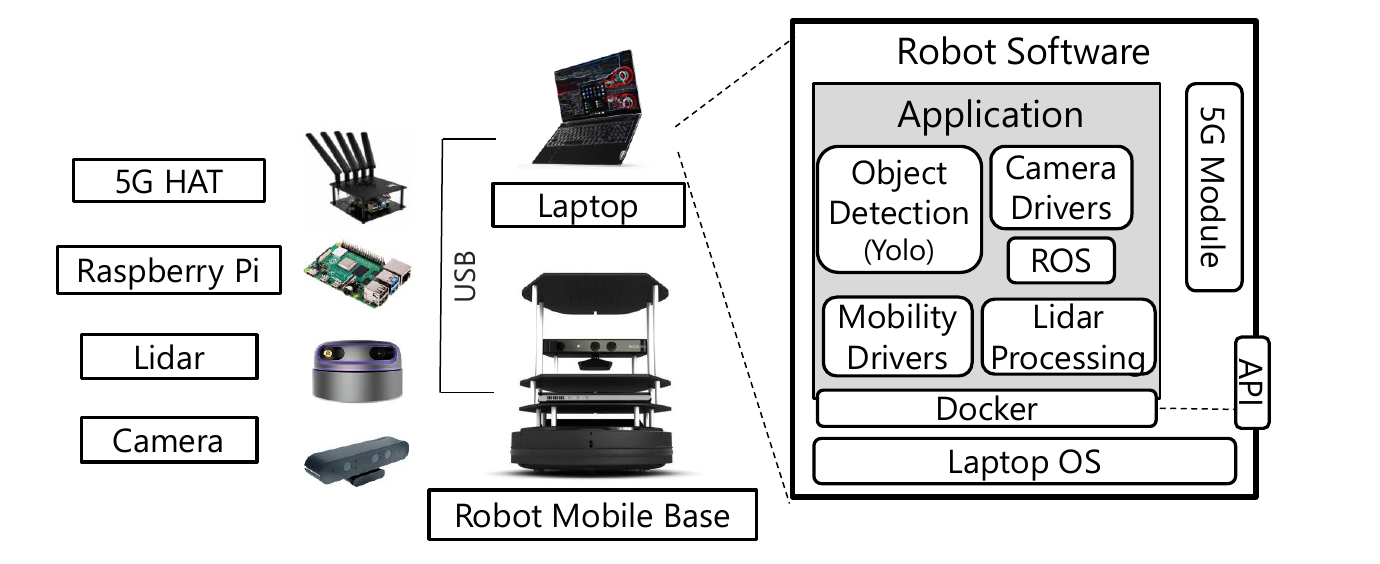}
    \caption{Robot software architecture.}
     \vspace{-3mm}
    \label{fig:robot_architecture}
\end{figure}

This section studies the energy consumption profile of the exemplary robotic application \revmin{using} ROS-compatible Kobuki TurtleBot2 S2 robots from the implemented testbed described in Sec.~\ref{sec:POC}. Each robot is equipped with a laptop that connects the robot mobile base, lidar, camera and Raspberry/5G HAT via USB, as shown in Fig.~\ref{fig:robot_architecture}. All the robot virtual functions that compose the robot application are hosted \revmin{on} the robot laptop and deployed as docker containers. While the lidar, camera and Raspberry/5G HAT use the USB for both power supply (via the \emph{laptop battery}) and data transmission, the robot mobile base uses the USB interface only for odometry data transmission. The robot base is powered independently by an embedded battery, referred to as the \emph{robot battery}, which supplies energy to the electric engine and wheels, enabling robot mobility. It is worth mentioning, that the scope of our study is to profile the computational energy consumption of the robotics software modules, focusing specifically on the laptop battery, which represents the energy bottleneck in our prototype. We leave for future work the study of the robot battery and the optimizations of the electric engine and
wheels. 

\subsection{Energy Consumption of Sensor Devices on the Robots}
\begin{figure}
    \centering
    \includegraphics[width=\columnwidth]{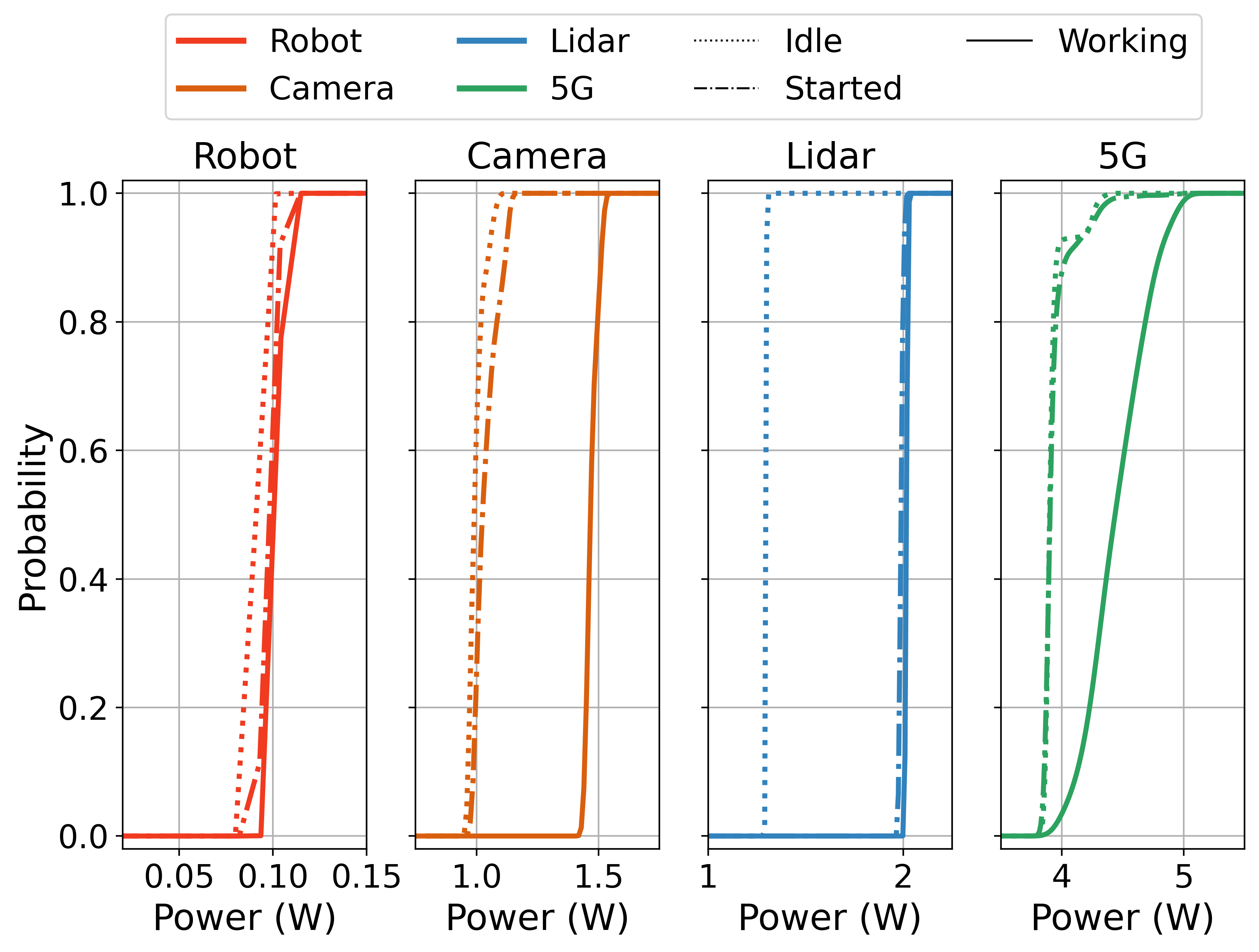}
    \caption{Power consumption of different hardware sensor devices reported by the USB meter.}
     \vspace{-2mm}
    \label{fig:usb_consumption}
\end{figure}

Our first experimental analysis aims at characterizing the power consumption of the different sensor and communication devices attached to the robot, as well as their impact on the \emph{laptop battery} when running their corresponding tasks. We adopt an UM34/UM34C USB meter to measure the power consumption of the different sensor devices that are powered by the \emph{laptop battery}. Each experiment runs for 30 minutes while collecting data from the USB meter every second. 
In particular, Fig.~\ref{fig:usb_consumption} shows the cumulative distribution function (CDF) of the power consumption differentiating for each sensor device and throughout different states, namely, \emph{Idle}, \emph{Started} and \emph{Working}.
The \emph{Idle} line refers to the power consumed when the device is in idle state, i.e., only connected via USB but not operating. Conversely, \emph{Started} line represents the power consumed by the device when USB-connected and having its corresponding ROS drivers active. Finally, the \emph{Working} line shows the device working state where the device is connected, its related ROS drivers are started, and its corresponding sensing data are generated, consumed and/or transmitted.     

From left to right, the first graph focuses on the power consumption of the robot drivers. We can notice low power consumption in all three states, mainly due to the USB connection being only used for data transmission, mainly related to odometry, robot state and navigation commands. We remark that the robot mobility is enabled by the \textit{robot battery}.
The second graph depicts the USB power consumption of the camera. Its values are approximately 1W when in \emph{Idle} or \emph{Started} state but increase up to 1.4W when entering the \emph{Working} state. This is because generally the USB port of the laptop is only used as a power source, while in the \emph{Working} state the camera starts a video stream which results in an increment of 0.4W.   
The third graph shows the USB power consumption of the lidar. When the device is in \emph{Idle}, its power consumption equals to~1.25W. This value goes up to 2W once the device starts\revmin{,} given that additional energy is required to spin the lidar at a given frequency and generate point cloud maps.
The last plot shows the USB power consumption of the 5G HAT. The power requested by the USB port for the \emph{Idle} state is the same as the one for \emph{Started} state, i.e., when the device is connected to the 5G network. This is natural since the 5G HAT requires a baseline power consumption to operate its circuitry. However, we can notice an increment of the power consumption when increasing the transmission rate over the 5G link, i.e., in \emph{Working} state, mainly due to modulation/demodulation processing.

\begin{figure}
    \centering
    \includegraphics[width=.9\columnwidth]{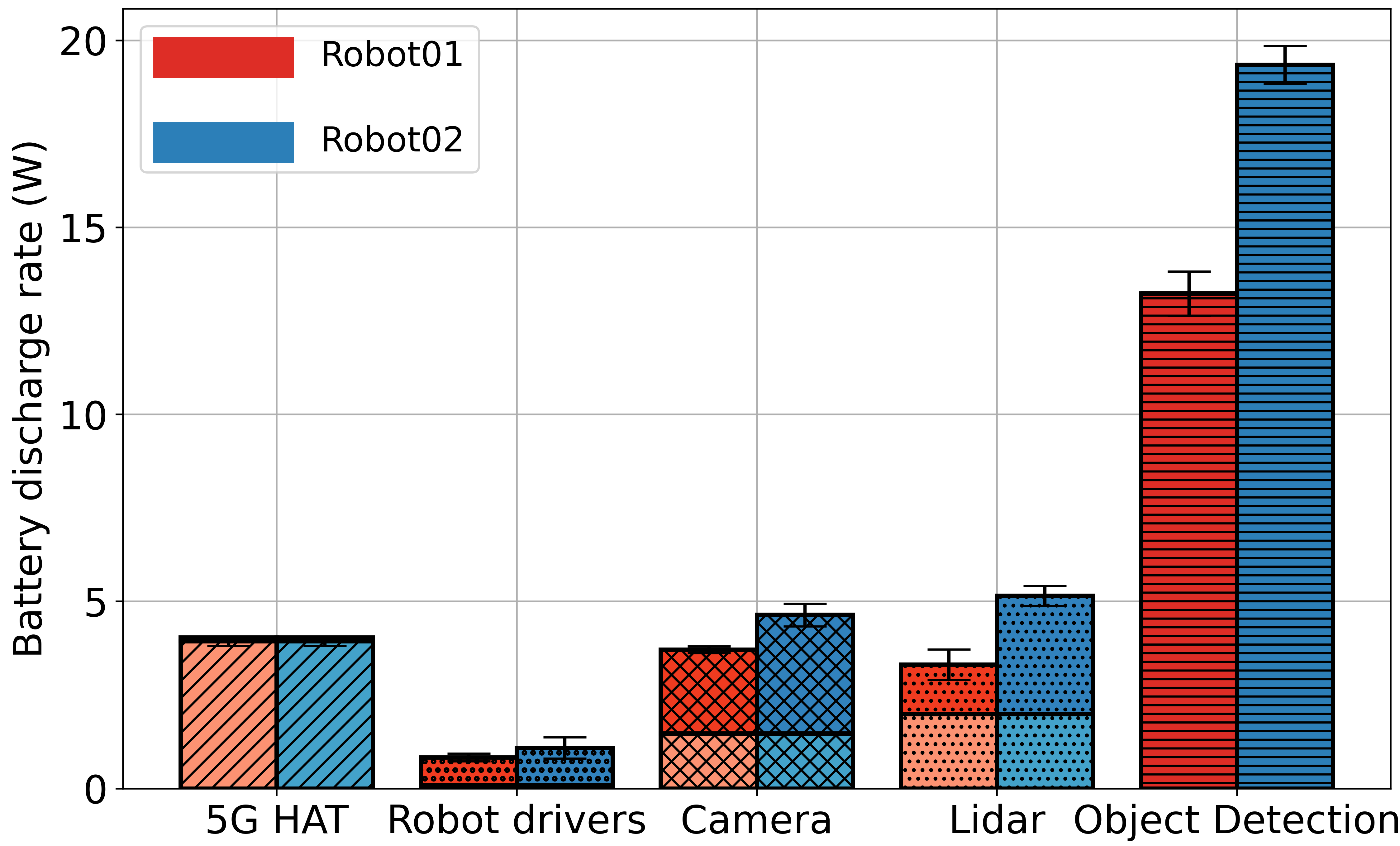}
    \caption{Battery discharge reported by the robots for different devices and ROS applications.}
     \vspace{-2mm}
    \label{fig:bat_discharge}
\end{figure}

\subsection{Energy Consumption of the Robot Virtual Functions}

In the following, we characterize the impact of the different software instances running as ROS modules and measure the total power consumption needed to run our robotic application.
The bar plot in Fig.~\ref{fig:bat_discharge} reports the energy consumption for \textit{i)} powering the different sensor devices on the robot and \textit{ii)} running the different ROS modules. It should be noted that the reported power consumption is the sum of the previous hardware-related experiments (light part of each bar), and the power required by the running software instances and/or their corresponding ROS nodes (dark part of each bar).
Since each robot has a different laptop characterized by \revmin{a} different processing power (Intel Core i5-1335U at 2.6 GHz, named as \textit{Robot01}, vs Intel Core i7-1250U at 4.7 GHz, named as \textit{Robot02}) and battery capacity (55 Wh (4-cell) vs 60 Wh (4-cell)), we detail the measurements for the two different robots.

From left to right, the first pair of bars shows the battery power consumption of the 5G HAT for the two robot laptops. They are both equal to the USB power consumption depicted before, as no local processing (modulation/demodulation) is required on the laptop to operate the 5G module for data plane communications.
The second pair of bars relates to the robot drivers. We can observe that the USB power consumption is rather small (e.g., 0.1 W), while most of the battery power consumption is due to the ROS nodes running on the laptop. A total of $5$ ROS nodes are necessary to control the robot and to manage its navigation, speed control and state publishing.
The third pair of bars shows the battery power consumption of the camera. In this case, in addition to the USB power requirements, a larger power consumption is necessary to run the $14$ ROS nodes managing the video streaming on the laptop. In fact, the 3D camera needs significant processing and thus energy to generate the video stream, re-configure camera depth metrics and perform real-time calibration.
The fourth pair of bars plot considers the lidar. Here, the dominant battery power consumption factor is the baseline power requested by the USB connection to spin the lidar at a given frequency while collecting point cloud data. A single ROS node manages the lidar drivers and exposes the scan information as a ROS topic. The generated sensing data is relatively light given that we adopt a 2D lidar capable of collecting about 1000 points spread over the $360^{\circ}$ area around the robot, hence the corresponding battery discharge is relatively low.
Finally, the object detection application presents the highest energy requirements, although composed by a single ROS node. In this case, the battery discharge is mainly due to the processing of the video stream, and the computing requirements of ML-based models that operate on each video frame in real-time to provide object detection and tracking. This is confirmed by a significant difference (5W) between the battery discharge of \textit{Robot01} and \textit{Robot02}. This is mainly due to the fact that the second robot in our deployment is equipped with a more powerful computing unit. As a result, the Neural Network of Yolo running on the second robot is more performant and consumes more processing resources, therefore resulting in an increased battery discharge rate.

Throughout the energy profiling of the various hardware and software composing our robotic application, we observe how each device poses different processing requirements which, in turn, affect the battery consumption in heterogeneous ways, depending on the type of sensors, their state, and their data transmission rate.
Therefore, in the following, we evaluate the potential benefits of remote robot sensor control in heterogeneous multi-robot scenarios.

\begin{figure}
    \centering  
    \vspace{-5 mm}
    \includegraphics[width=\columnwidth]{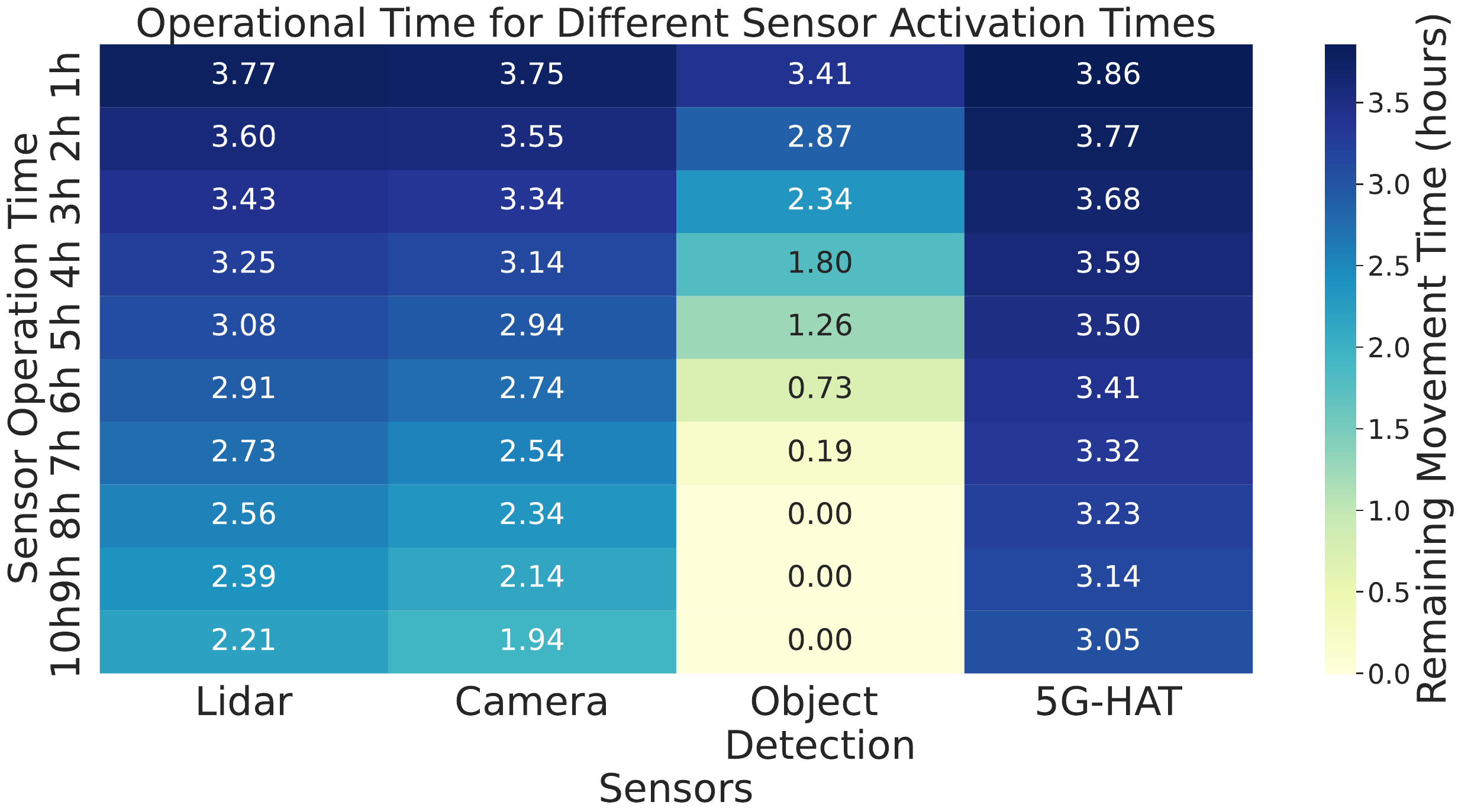}
    \caption{\rev{Heatmap of the robot remaining movement time for different sensor activation times.}}
    \label{fig:sensor_control}
\end{figure}

\subsection{\rev{Robot sensor control}}

\rev{Fig.~\ref{fig:sensor_control} illustrates the relationship between the operational time of different sensors/ROS applications and the remaining movement time of the robot, revealing distinct energy consumption patterns. Overall, the remaining movement time of the robot decreases as sensors and ROS application operation time increases. Among the ROS applications, the \revmin{o}bject detection (Yolo) exhibits the steepest decline, with the remaining movement time dropping to zero by 8 hours, indicating its high energy consumption. In contrast, the 5G \revmin{H}AT consistently maintains the highest remaining movement time, exceeding 3 hours even at 10 hours of operation, demonstrating superior energy efficiency. The \revmin{l}idar and \revmin{c}amera sensors show a gradual and comparable decline in remaining movement time, with the \revmin{l}idar retaining slightly more time (2.21 hours) than the \revmin{c}amera (1.94 hours) at 10 hours of operation. At shorter operation times, the differences among all the components are minimal, with remaining movement times ranging between 3.41 and 3.86 hours. However, as the operation time increases, the disparities become more pronounced, highlighting significant differences in energy efficiency. These observations suggest that the smart control of the \revmin{o}bject detection ROS application together with the camera and \revmin{l}idar sensors have the potential to significantly reduce the system energy consumption, particularly during prolonged operations. On the other hand, the 5G \revmin{H}AT is well-suited for extended use, making it a favorable choice for energy-constrained networked robotics scenarios.}
\begin{figure*}[t!]
    \centering
    \begin{subfigure}[t]{.3\textwidth}
    		\centering
        \includegraphics[trim = 0cm 0cm 0cm 0cm, clip, width=\columnwidth]{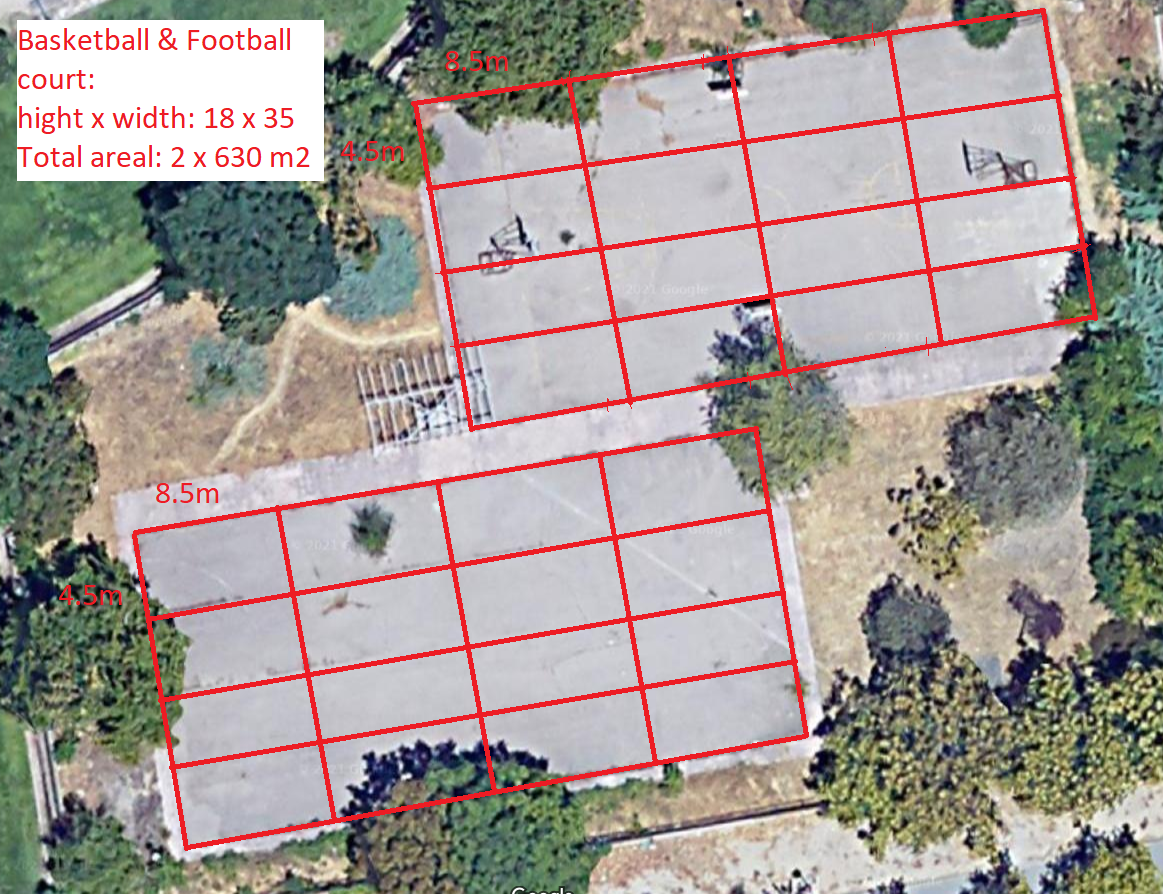}
        \caption{\small Top view of outdoor experimental area.}
        \label{fig:testbed_top_view}
    \end{subfigure}
    \begin{subfigure}[t]{.34\textwidth}
    		\centering
        \includegraphics[trim = 6cm 0cm 6cm 0cm, clip,width=\columnwidth]{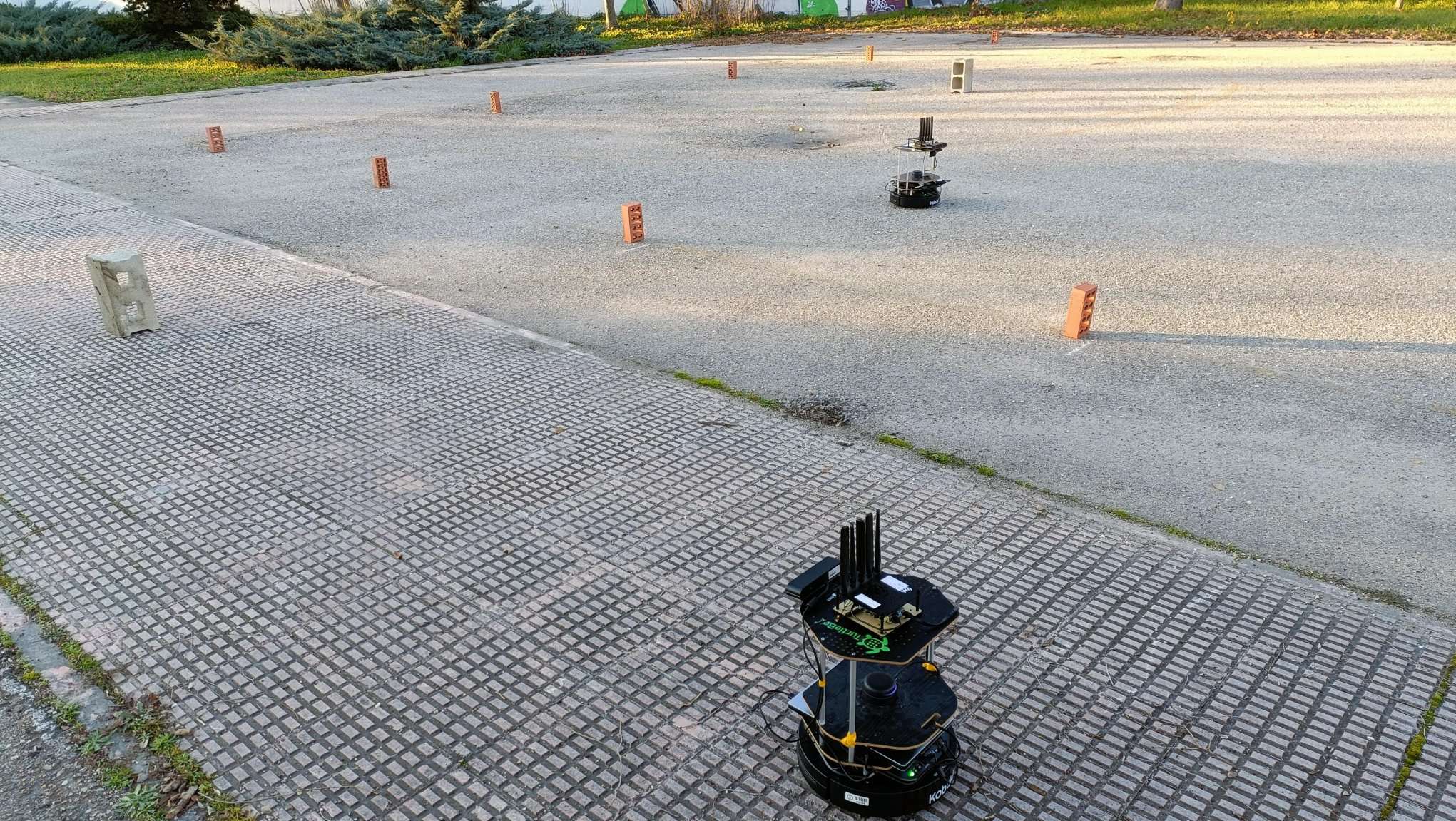}
        \caption{\small 5G-Enabled Turtlebot and obstacles.}
        \label{fig:testbed_robots}
    \end{subfigure}
    \vspace{2mm}
    \caption{Experimental testbed architecture and scenario.}
    \label{fig:real_scenario}
\end{figure*}

\section{Field Test Validation}
\label{sec:outdoor}
This section presents the experimental results from the outdoor field tests that we performed using our PoC implementation of the \emph{\name{}} system (described in Sec.~\ref{sec:POC}) in the campus of \revmin{the} Universidad Carlos III de Madrid in Spain, using the 5G infrastructure provided by 5TONIC~\cite{5TONIC}. Fig.~\ref{fig:testbed_top_view} depicts the outdoor area considered in our experiments, while Fig.~\ref{fig:testbed_robots} depicts the robots and the ground obstacles placed into the target zone. The overall area is logically split into grid elements, following the description of Sec.~\ref{subsec:\name{}} and the corresponding parameters defined in Table~\ref{tab:experimentalsetup}. The sports pitches are covered by a RAN infrastructure composed of a single 5G base station configured with 40 MHz channels at a carrier frequency of 3500 MHz (5G band n78) to support the communication between the robots and the edge datacenter. 

\begin{table}
\centering
\vspace{5mm}
\caption{Experimental Setup}
\begin{tabular}{l|l||l|l}
\hline
\textbf{Definition} & \textbf{\emph{\name{}} Value} & \textbf{Definition} & \textbf{\emph{\name{}} Value} \\
\hline
$|\mathcal{T}|$     & 35                      &    $|\mathcal{R}|$     & 2                 \\
$A \times B$        & $13\times9$   subareas    &  $|g_{a,b}|$         & $3.7\times3.1$ $m^2$    \\
%$B_{max}$           & \tbd{$5000$ J    }            &   
%$v$                 & \tbd{$1$, $\sqrt{2}$ m/s }   \\
\hline
\end{tabular}
\label{tab:experimentalsetup}
\end{table}

The experimental drive started with both robots streaming the sensor data (i.e., robot position, battery status) via the 5G HAT to the edge datacenter where \emph{\name{}} is deployed. Upon receiving the real-time information from the robots, \emph{\name{}} computes the next position and decides the sensors state (ON or OFF) for each robot. The subsequent navigation goal command is sent from \emph{\name{}} to the Robot Orchestrator that, based on the current localization of the robots, navigates both robots in a coordinated closed-loop manner. The sensors state command is sent from \emph{\name{}} to the 5G Orchestrator, which in turn decides if the sensors need to be turned ON or OFF depending on the real-time status of the system. During the complete discovery of the outdoor area, we measure the 
\revmin{d}ocker-related statistics in both the robots and the edge (CPU consumption, RAM consumption and Tx/Rx data) for all the virtual application functions of our system. In addition, we measure the battery status in the robots, battery discharge and the robot odometry. 

\begin{figure}[t]
    \centering
\includegraphics[width=\columnwidth]{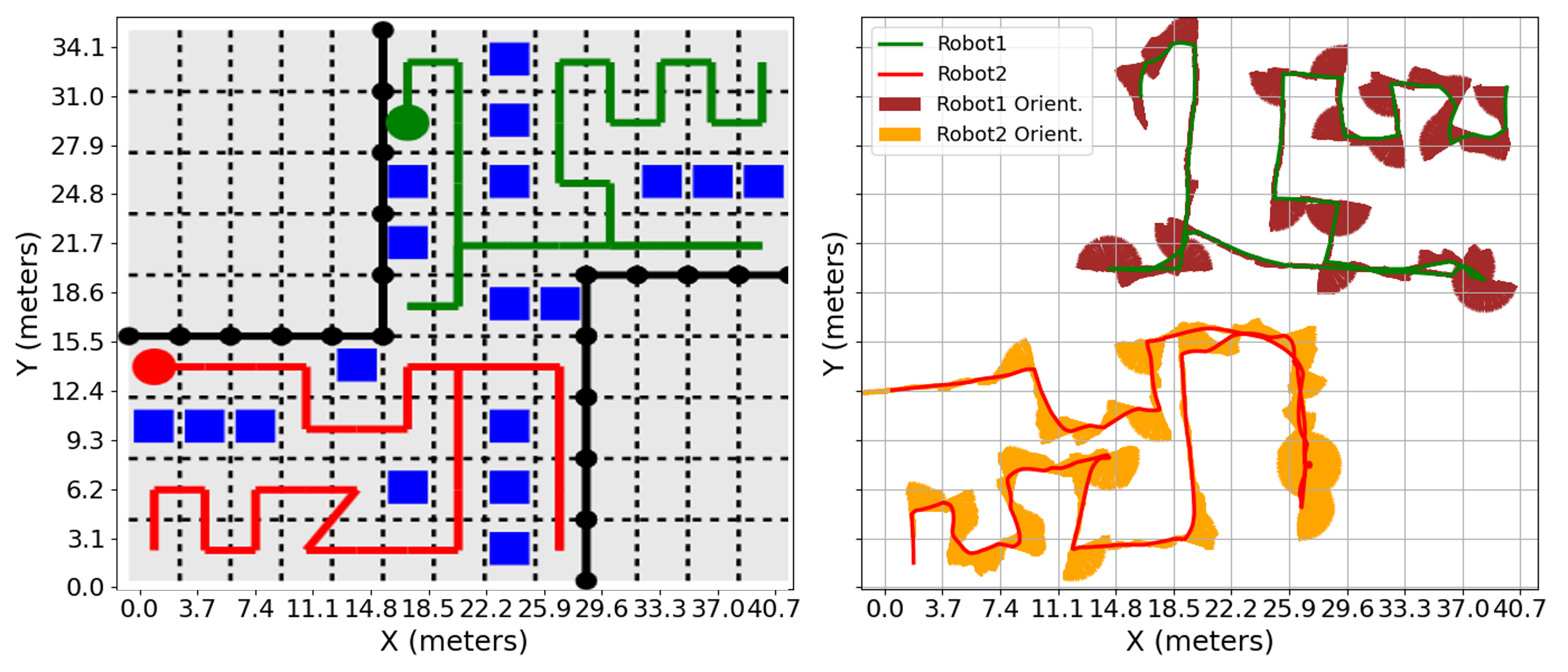}
    \caption{Comparison of theoretical (left) and executed robot trajectories (right) with yaw orientation angle.}
     \vspace{-2mm}
    \label{fig:trajectories}
\end{figure}

To evaluate the performance of our solution and its impact on \revmin{the} robot KPIs like CPU and energy consumption, we consider the following benchmark, dubbed as State-of-the-Art (SOA). In the SOA case, \emph{\name{}} is not used in the exploration task optimization and sensor control, and neither edge platform is used. That means, robots running with SOA settings can not offload any computational tasks to the edge, and thus they need to run every sensor and related processing locally. For the same reason, the different processing functions are statically deployed in the robots. This provides the baseline of our experiments and allows us to characterize the energy consumption of the devices in this unknown environment. In the SOA scenario, we deploy the robots in the experimental area and let them move and explore the area by sensing the environment until reaching full discovery. To ensure a fair trajectory comparison, we impose the same navigation path when instantiating \emph{\name{}} and SOA scenarios. We remark that the path is dynamically extracted from the algorithm upon detection of obstacles in the \emph{\name{}} case, and re-used in \revmin{the} SOA approach.
The discovery phase of our outdoor area took about $10$ minutes. The left graph in Fig.~\ref{fig:trajectories} depicts the theoretical optimum trajectories (calculated a posteriori by our simulator, as presented in Sec.~\ref{subsec:simulator}), following the real-time robot odometry collected along the exploration phase. It can be noticed how the trajectories performed by the two robots follow the obstacle distribution, i.e., robots successfully detect the obstacles in their proximity and report to \emph{\name{}}. The right graph shows the executed trajectories of the robots in the experiments, which are not perfectly aligned with the theoretical ones. 
This limitation arises because the precision of the robot’s trajectories relative to the theoretical optimum is constrained by the hardware capabilities, including the accuracy of odometry measurements, the reporting system, and the motor control frequency of the robot.

\begin{figure}
    \centering
\includegraphics[clip, trim = 0.5cm 0cm 1cm 0.5cm, width=\columnwidth]{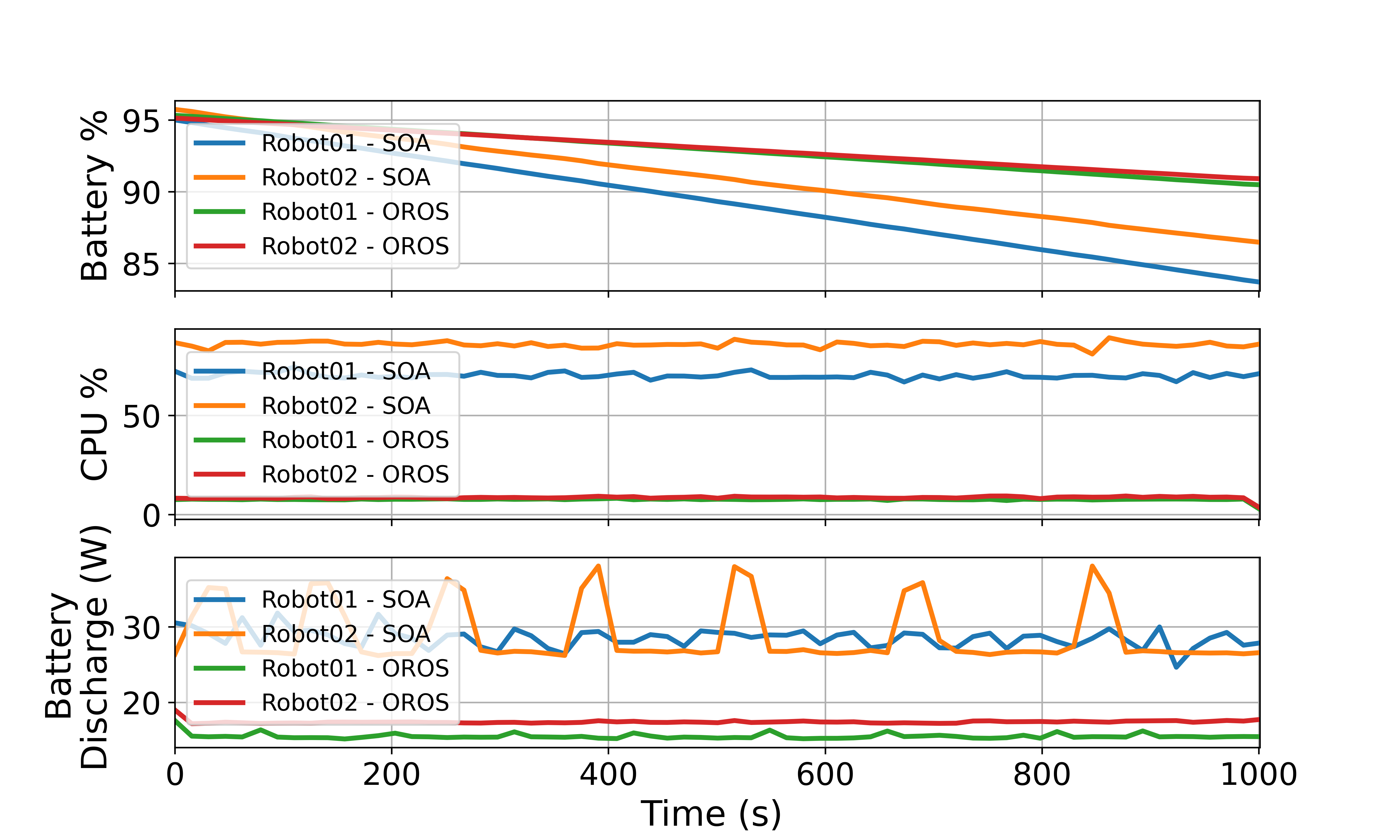}
    \caption{\rev{Robot battery usage and CPU consumption over time.}}
   \vspace{-2mm}
\label{fig:robot_stats}
\end{figure}

\begin{figure}[t]
    \centering
    \includegraphics[width=\columnwidth]{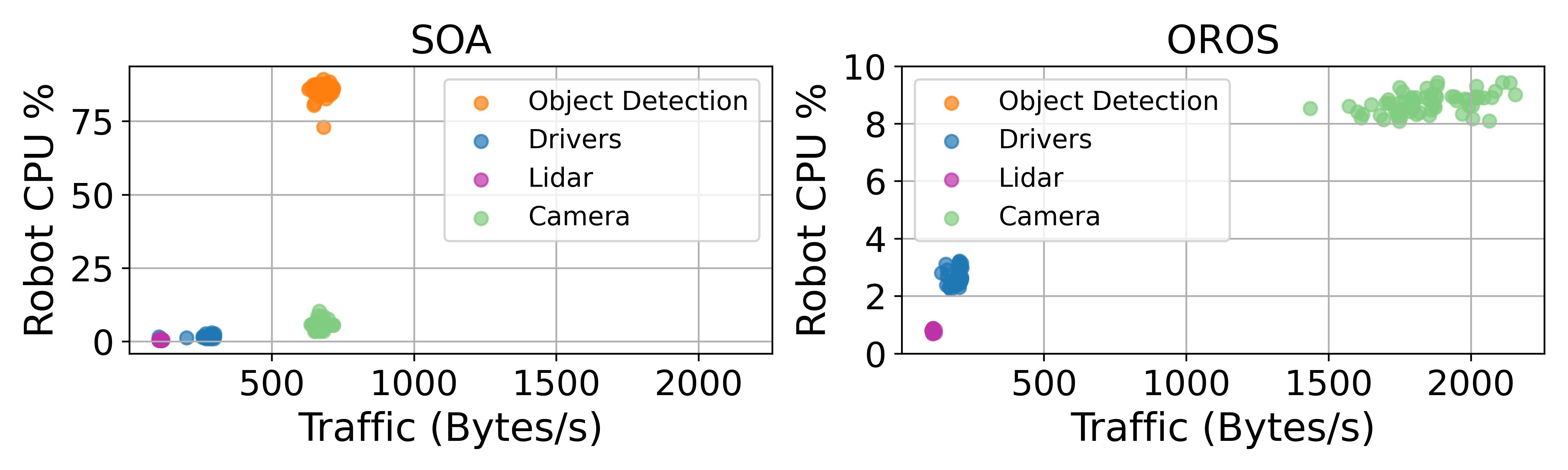}
    \caption{\rev{Robot CPU utilization as a function of generated traffic load.}}
    \label{fig:robot_cpu}
    \vspace{-2mm}
\end{figure}

Fig.~\ref{fig:robot_stats} depicts the battery percentage, CPU utilization, as well as the instantaneous discharge rate on the robots over time for the two deployed robots. In the middle plot, we can observe how SOA settings render to $\sim70\%$ higher CPU load compared to the \emph{\name{}} settings for both robots. This is mainly due to the heavy local processing of the object detection virtual function which, in turn, impacts the potential operational time. The high CPU load translates to a faster decrease of the battery percentage on both robots, as illustrated in the top graph. It is worth mentioning that the small difference in the resulting battery level of the two robots at the end of the same experimental settings (i.e., SOA and \emph{\name{}}) is mainly due to different laptop models installed on the robots, each one with its own maximum battery size and CPU model, which impact the corresponding processing capabilities. This is also shown in the bottom graph, where we can identify heterogeneous behavior in the discharge rates, especially in the SOA settings when compared to \emph{\name{}}.

\begin{figure}
    \centering
    \includegraphics[width=\columnwidth]{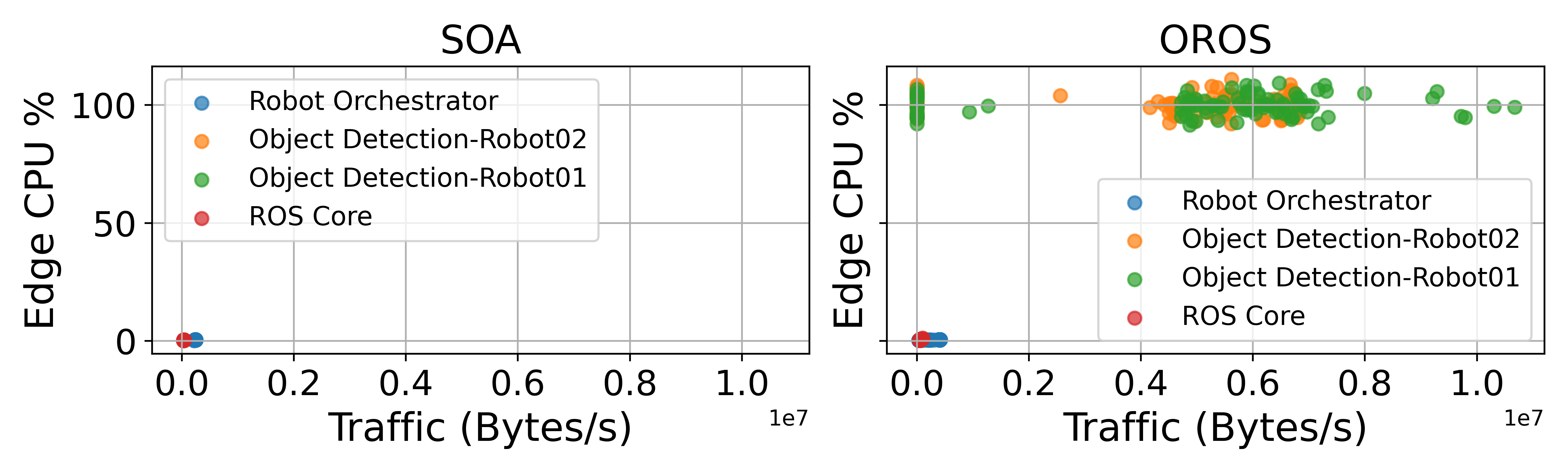}
    \caption{\rev{Edge CPU utilization and generated and received traffic per running function.}}
     \vspace{-2mm}
    \label{fig:edge_stats}
\end{figure}

\rev{Fig.~\ref{fig:robot_cpu} and Fig.~\ref{fig:edge_stats} illustrate the CPU utilization of the robot and the edge platform, respectively, as a function of the generated traffic volume over 1 second time interval from different virtual functions in the system.
In both figures, the graphs on the left depict the CPU utilization as a function of the traffic in the SOA experimental scenario, while the graphs on the right depict the CPU utilization as a function of traffic in the \textit{OROS} experimental scenario. In Fig.~\ref{fig:robot_cpu}, it appears how different functions pose different computing loads into the robot, with \revmin{o}bject \revmin{d}etection traffic leading around 80\% of CPU utilization at about 650 Bytes/s while \revmin{d}rivers, \revmin{l}idar and \revmin{c}amera maintaining very low CPU usage (below 8\%). Similar behaviors were also identified by previous results shown in Fig.~\ref{fig:bat_discharge}, suggesting the adoption of an edge computing platform where to offload such heavy processing. Conversely, in the \textit{OROS} scenario, it can be noticed how larger traffic volume flowing through the radio interface, mainly related to camera streaming, actually has a limited impact on the robot CPU utilization $\leq10\%$.
Similarly, the CPU usage related to \revmin{d}rivers and \revmin{l}idar remains very low, while object detection is not present due to the \textit{OROS} scenario's edge offloading policy.

In contrast, for the CPU utilization in the edge, the data from Fig.~\ref{fig:edge_stats} illustrates that for the traffic of the SOA scenario, the edge platform experiences very low CPU utilization (under 5\%) with limited \revmin{R}x and \revmin{T}x packet volumes (up to $2500$ KBytes/s) for the Robot Orchestrator and ROS \revmin{c}ore components. The \textit{OROS} scenario, as a result of the offloading policy, shows larger CPU utilization (up to $100$\%) in the edge for the \revmin{o}bject \revmin{d}etection on both robots, handling much larger packet volumes (up to $10$ MBytes/s).}

\begin{figure}
    \centering
    \includegraphics[width=\columnwidth]{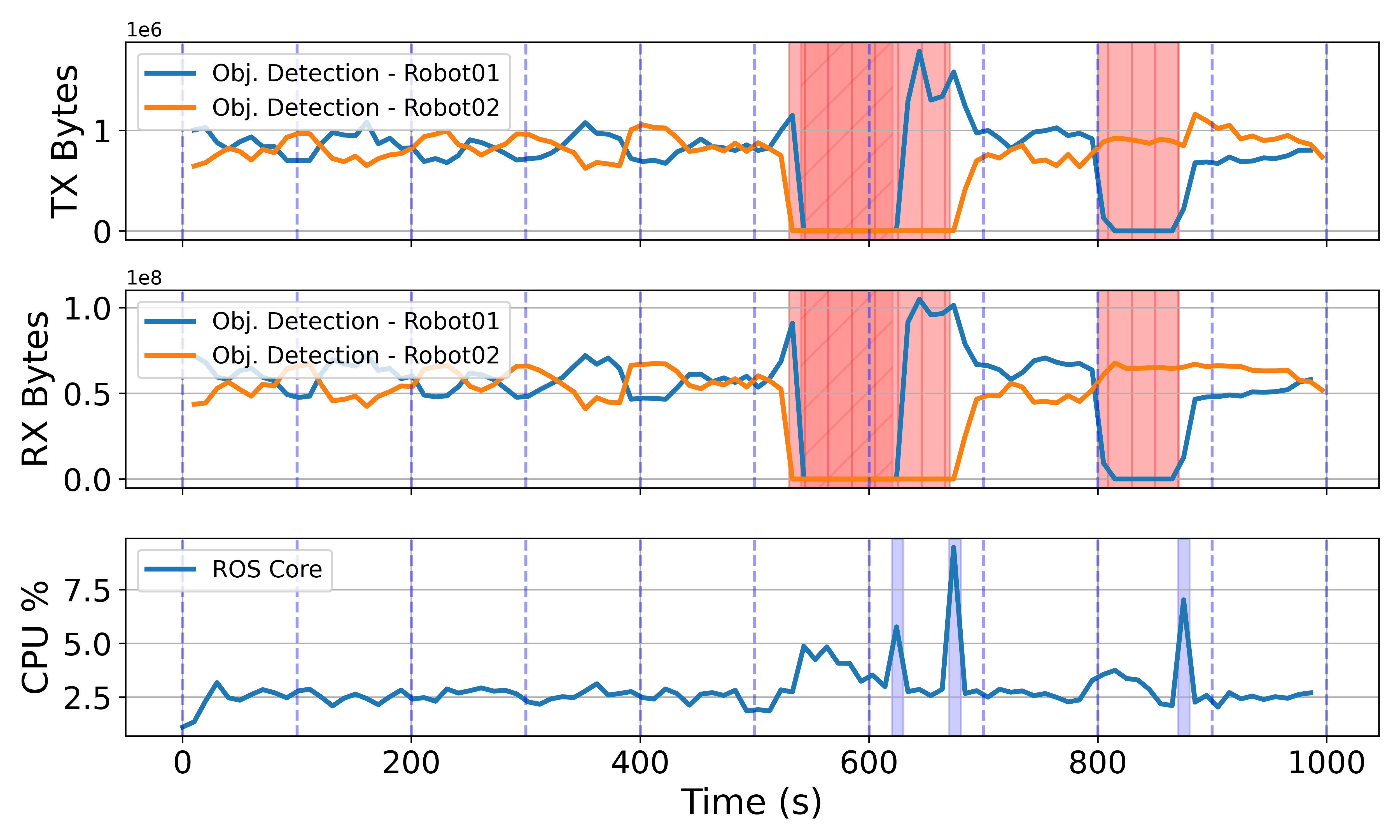}
    \caption{ROS core ontainer CPU utilization and generated and received traffic over time.}
     \vspace{-2mm}
    \label{fig:experiment}
\end{figure}

Fig.~\ref{fig:experiment} characterizes the impact of \emph{\name{}} when making dynamic orchestration decisions in our multi-robot deployment, focusing on the object detection traffic generated by individual robots and the corresponding effect on the edge platform over a subset of decision intervals.
In particular, the highlighted red area highlights the time periods when \emph{OROS} imposes the decision to switch off the robot’s sensors. In the same plot, dashed blue lines identify the occurrence of a new decision interval set to 100 seconds. From the plots, it can be noticed how robots switched off their camera sensors when going through an already explored area element which, in turn, stops the traffic towards the edge platform. The traces also highlight a small delay (in the order of few seconds) between the command transmission from \emph{\name{}} and its effective execution at the robot side. This is mainly due to the multiple entities involved in our prototype at both ROS and 
\revmin{d}ocker levels. Moreover, we can notice in the bottom figure how the re-activation of the sensors in subsequent time intervals causes a significant peak in the CPU utilization of the ROS \revmin{c}ore container. This is because each sensor activation triggers tens of ROS nodes (that are part of the camera, lidar and object detection virtual functions) to register and/or subscribe to different ROS topics as well as services in the ROS core. For example, the camera sensor alone needs to register approximately 8 ROS topics and 17 ROS services\footnote{\label{camera-ros} https://wiki.ros.org/astra\_camera}. 

As a summary, in the field test trials we successfully moved \emph{\name{}} from \revmin{a} simulation environment to a real-world testbed. In the experimental evaluation, not only we validated the feasibility of using \emph{\name{}} to jointly manage multi-\revmin{r}obots and a local computing infrastructure, but also we demonstrated the  significant energy savings gains achieved by offloading demanding computational tasks to the 5G edge infrastructure and dynamic energy management of on-board sensors (e.g., switching them off when they are not needed). This strategy achieves approximately $\sim70\%$ lower CPU load and in turn $\sim15\%$ energy savings on the robots, thereby extending the robot battery life, allowing for longer operating times.

\section{Scalability: Algorithm Complexity and Potential Solutions}
\label{sec:scalability}
\rev{
Scalability in real-time decision-making systems is heavily influenced by algorithmic complexity and the efficient utilization of computational resources. This becomes particularly relevant in dynamic and resource-constrained environments, where multiple robots operate over large exploration areas. In the following, we analyze the scalability challenges of \textit{\name{}}, focusing on algorithm complexity, and discuss potential solutions to address them.

As discussed in Sec.~\ref{sec:framework}, \textit{\name{}} leverages a decision-making model operating within a configurable time window, \( \mathcal{W} = \{ t, \dots, t+W \} \subseteq \mathcal{T} \), where \( W \) represents the size of the decision window. This method enables the system to scale for real-time operations by prioritizing short-term planning and adaptability. Due to the adoption of binary decision variables however, the worst-case computational complexity of the MILP model is given by $O(2^{|r| \cdot |W| \cdot |A| \cdot |B|})$, where \( |r| \) is the number of robots, \( |W| \) is the size of the decision window, and \( |A| \) $\cdot$ \( |B| \) represent the number of subareas in the deployment environment. 
This exponential complexity highlights significant challenges when scaling to larger environments or higher numbers of robots. Worst-case scenarios, such as deployments at the boundaries of exploration areas without overlapping trajectories, exacerbate these demands, necessitating further optimization techniques.  

To address these scalability challenges, several strategies can be employed. Problem decomposition is an effective technique, where the MILP problem is divided into smaller sub-problems focused on the local exploration areas of each robot. Advanced optimization techniques such as column generation or Benders decomposition can further enhance this process by iteratively solving smaller sub-problems while maintaining global optimization. Additionally, search space reduction can be achieved through methods such as branch-and-bound or branch-and-cut, which eliminate large portions of infeasible solutions, thus improving computational efficiency.

Another approach involves dynamically adjusting the decision window size, \( W \), along the exploration path. This allows the system to adapt to varying levels of complexity in different regions, striking a balance between computational efficiency and decision accuracy. In regions with higher complexity, the decision window can be narrowed, whereas in simpler areas, it can be expanded to improve planning horizons without overwhelming computational resources. Furthermore, leveraging context-awareness in environments with dense obstacles can reduce computational demands, as the restricted movement options for robots simplify the problem space. 

Empirical evaluations of the \textit{OROS} framework available in~\cite{OROS_TNSM} have provided insights about its scalability and performance. The average computational solver time was observed to remain within a few seconds, even as the number of robots and window size \( W \) increased, demonstrating its feasibility for real-time deployments. Interestingly, the presence of more obstacles reduced computational solver time due to the constrained movement options available to robots. To exploit this fact, robots may be deployed and start the exploration from nearby locations.
These results underscore the importance of context-awareness in enhancing scalability. Despite the initial PoC experiments being limited to two robots due to testbed constraints, future evolution of the testbed can integrate advanced orchestrators for larger-scale deployments, enabling a comprehensive evaluation of scalability.

The integration of advanced optimization techniques, dynamic decision window adjustments, and robust orchestration frameworks will further enhance scalability. Future work will explore these aspects, ensuring that the system can handle larger-scale deployments while maintaining computational efficiency and decision accuracy. The dynamic adaptation of the decision window and the use of advanced optimization techniques will play a key role in supporting dynamic and large-scale robotic deployments in real-world scenarios.
}
\section{Related Work}
\label{sec:related}
\vspace{-1 mm}

\subsection{\rev{Robotic Orchestration}}

\rev{The field of robotics has witnessed significant advancements, with emerging academic and industrial platforms like FogROS2~\cite{fogros}, RobotKube~\cite{robotkube}, KubeROS~\cite{kuberos}, NVIDIA Isaac Sim~\cite{isaac-sim}, Open-RMF~\cite{OPENRMF}, and AWS RoboMaker~\cite{robomaker} leveraging cloud and edge computing to orchestrate robotic services. Robotic platforms like, FogROS2, RobotKube and KubeROS are Kubernetes-based frameworks that automate the orchestration and management of virtualized robotic applications in heterogeneous computing infrastructures. However, the goal of these platforms is widespread adaptation of edge and cloud computing in robotic systems and not the energy efficient joint orchestration of the 5G and robotic domains.} 

\rev{In the context of dynamic robot orchestration in SAR missions, the authors in~\cite{dynorch1} propose an orchestrator that makes its deployment decisions based on specific parameters (e.g., required RAM, GPU) and adapts to changes in these factors dynamically, making the system able to react to external influences. Similarly, the authors in~\cite{dynorch2} present a Kubernetes orchestrator that provides flexible and scalable life-cycle management of UAV Rescue Operations service. While these works are an example of dynamic orchestration of SAR use cases, they only manage the virtual services and the underlying infrastructure resources. They do not try to optimize the robot path planning nor the energy consumption of the robots. From the industrial initiatives, platforms like NVIDIA Omniverse, Open-RMF and AWS RoboMaker, are focused on developing, testing, and deploying intelligent robotics applications, overlooking completely the 5G domain.}

\subsection{\rev{5G Orchestration}}

\rev{The 5G domain has a long-lasting interest in supporting robotic applications from a networking perspective, particularly focusing on how 5G and beyond 5G architectures can support these time-sensitive applications. While several studies have explored the use of 5G for robotic control and remote operation, such as~\cite{5g-surgery} on 5G-enabled remote robotic surgery,~\cite{5g-robotics} on 5G-enabled mobile robots, and~\cite{robot-dt} on 5G-enabled robotic digital twins, these studies often focus on specific use cases and feasibility, and do not address the broader challenges of jointly orchestrating the 5G and robotic domains. For instance, while~\cite{5g-surgery} delve into the technical aspects of remote surgery, they do not consider the energy efficiency and resource management challenges associated with robotic operations. Similarly,~\cite{robot-dt} explores the benefits of 5G and robot function virtualization for robot manipulators, but does not provide a comprehensive framework for dynamic resource allocation and task navigation in multi-robot scenarios. }

The authors of~\cite{Voigtlnder2017} propose a framework that offloads time-critical and computationally intensive tasks onto a distributed node architecture, leveraging 5G communication between robots and cloud servers to enhance operational efficiency. However, the limited energy available from on-board batteries remains a significant challenge in practical environments. While offloading computational tasks can conserve energy, robots still face a trade-off between battery size and energy consumption. Larger batteries extend mobility but increase overall energy consumption due to their weight, as discussed by Albonico \textit{et al.}~\cite{Albonico2021}.
Swanborn \textit{et al.}~\cite{Swanborn2020} identify navigation as the primary energy consumer in robotic operations. They also highlight secondary sources of energy consumption, including inefficient hardware, poor management algorithms, idle times, operational inefficiencies (e.g., poor-quality software causing unnecessary stops and sharp accelerations/decelerations), processing energy, and excessive communication and sensor data acquisition. Addressing these inefficiencies is crucial for improving energy efficiency in robotic systems.

\subsection{\rev{Dynamic discovery}}
\rev{When it comes down to optimizing the discovery of the unknown areas, the authors in~\cite{dynamic-task1} propose a groping algorithm that reduces the robot-to-robot communication loss rate while improving the robot task execution efficiency. In this context, the authors in~\cite{dynamic-task2} propose a framework that optimizes the multi-robot task allocation based on robot capabilities, victim requirements, and past robot performance. These works are an example of dynamic robot task allocation, however they neglect the need for optimization of the 5G domain and robot energy efficiency.} To fill in the energy efficiency gap, authors in~\cite{Rappaport2016} present an energy-efficient path-planning approach for autonomous mobile robots, minimizing the overall travel distance to reduce energy consumption. Motivated by the need for an energy efficient joint orchestration of the robotics and networking domain, we proposed the initial ideas of \emph{\name{}} in~\cite{OROS_WowMom} to connect both the orchestration entity from the network and the robot domains, enabling interaction and information exchange between the robots and the network infrastructure. Based on the offline optimization model formulated in~\cite{OROS_WowMom}, we further developed a heuristic online approach in~\cite{OROS_TNSM} that is more suitable for real-time robot discovery operations.
However, while the study conducted in both works led to promising results, they derive from pure simulation environments and lack feasibility implementation and validation in real-world scenarios.
Therefore, it becomes evident the need for a comprehensive field testing to fully evaluate the framework's effectiveness and practicability, addressing the gaps in implementing the required APIs to interconnect the robotic and 5G domains, as discussed in this paper.

\section{Conclusions \rev{and Future Work}}
\label{sec:conclusion} 

The advent of ubiquitous and low-latency communication provided by 5G networks paved the road for collaborative robotic use cases leveraging a flexible edge/cloud infrastructure for data sharing and processing of cloud-native robotic applications.
Robot operating systems however were designed as closed systems, not inherently built to communicate with external platforms, causing robots to perform all tasks locally. This lack of integration leads to high energy consumption, negatively impacting operational efficiency.
To overcome this gap, we developed an energy-efficient joint orchestration solution to interconnect the 5G and the robotic domains. \emph{\name{}} considers both robot and communication infrastructure monitoring information to jointly determine the optimal robot navigation strategy and the best cloud-computing resource allocation, which, in turn, minimizes energy consumption and extends the robot exploration range.
We validated \emph{\name{}} in a real-world testbed exploiting commercial off-the-shelves robot devices, heterogeneous sensor hardware, and a fully-fledged 5G standalone mobile network. The experimental results show a significant gain of \emph{\name{}} for collaborative robot operations by reducing $\sim70\%$ CPU load and in turn $\sim15\%$ energy savings on the robots, providing a future-proof sustainable solution for emerging 5G-enabled robotic applications.

\rev{Future work will include field tests in larger environments with a growing number of heterogeneous robots and sensors. To address computational complexity in these scenarios, we plan to explore and implement heuristic and/or ML-based approaches, which can provide scalable and efficient solutions while maintaining acceptable performance. Moreover, the \revmin{\textit{\name{}}} orchestration framework can be further advanced to perform dynamic offloading decisions and derive sensor control policies in dynamic network conditions and varying environments.}

\section*{Acknowledgment}
The research leading to these results has been supported in part by the CERCA Programme / Generalitat de Catalunya, by the European Union’s H2020 6GGOALS Project (grant no. 101139232), by the SNS JU under the European Union's Horizon Europe MultiX Project (grant no. 101192521) and Predict Project  (grant no. 101095890), and by the Spanish Ministry of Economic Affairs and Digital Transformation and the European Union – NextGeneration EU (Call UNICO I+D 5G 2021, ref. number TSI-063000-2021-6 and TSI-063000-2021-122).

\vspace{-10pt}

\bibliographystyle{IEEEtran}
\bibliography{main}

\begin{thebibliography}{10}
\providecommand{\url}[1]{#1}
\csname url@samestyle\endcsname
\providecommand{\newblock}{\relax}
\providecommand{\bibinfo}[2]{#2}
\providecommand{\BIBentrySTDinterwordspacing}{\spaceskip=0pt\relax}
\providecommand{\BIBentryALTinterwordstretchfactor}{4}
\providecommand{\BIBentryALTinterwordspacing}{\spaceskip=\fontdimen2\font plus
\BIBentryALTinterwordstretchfactor\fontdimen3\font minus \fontdimen4\font\relax}
\providecommand{\BIBforeignlanguage}[2]{{%
\expandafter\ifx\csname l@#1\endcsname\relax
\typeout{** WARNING: IEEEtran.bst: No hyphenation pattern has been}%
\typeout{** loaded for the language `#1'. Using the pattern for}%
\typeout{** the default language instead.}%
\else
\language=\csname l@#1\endcsname
\fi
#2}}
\providecommand{\BIBdecl}{\relax}
\BIBdecl

\bibitem{rs15133266}
\BIBentryALTinterwordspacing
M.~Lyu, Y.~Zhao, C.~Huang, and H.~Huang, ``Unmanned aerial vehicles for search and rescue: A survey,'' \emph{Remote Sensing}, vol.~15, no.~13, 2023. [Online]. Available: \url{https://www.mdpi.com/2072-4292/15/13/3266}
\BIBentrySTDinterwordspacing

\bibitem{drones7050322}
\BIBentryALTinterwordspacing
N.~I. Sarkar and S.~Gul, ``Artificial intelligence-based autonomous uav networks: A survey,'' \emph{Drones}, vol.~7, no.~5, 2023. [Online]. Available: \url{https://www.mdpi.com/2504-446X/7/5/322}
\BIBentrySTDinterwordspacing

\bibitem{5GvsLTE}
M.~Aleksy, F.~Dai, N.~Enayati, P.~Rost, and G.~Pocovi, ``{Utilizing 5G in Industrial Robotic Applications},'' in \emph{International Conference on Future Internet of Things and Cloud (FiCloud)}, Aug. 2019, pp. 278--284.

\bibitem{5g_private}
\relax{Han Sr \emph{et al.}}, ``5g key technologies for helicopter aviation medical rescue,'' \emph{J Med Internet Res}, vol.~26, p. e50355, Aug 2024.

\bibitem{5g_private_1}
\relax{Han Wei Ji \emph{et al.}}, ``{Exploring innovative applications of 5G in aviation medical rescue},'' \emph{Hong Kong Journal of Emergency Medicine}, vol.~31, no.~2, pp. 84--88, 2024.

\bibitem{Swanborn2020}
S.~Swanborn and I.~Malavolta, ``{Energy Efficiency in Robotics Software: A Systematic Literature Review},'' \emph{IEEE/ACM International Conference on Automated Software Engineering Workshops (ASEW)}, pp. 144--151, 2020.

\bibitem{HIRO_NET}
L.~Ferranti, S.~D'Oro, L.~Bonati, E.~Demirors, F.~Cuomo, and T.~Melodia, ``{HIRO-NET: Self-Organized Robotic Mesh Networking for Internet Sharing in Disaster Scenarios},'' in \emph{IEEE International Symposium on "A World of Wireless, Mobile and Multimedia Networks" (WoWMoM)}, 2019, pp. 1--9.

\bibitem{Mission_Critical_Edge}
P.~Skarin, W.~Tärneberg, K.-E. Årzen, and M.~Kihl, ``{Towards Mission-Critical Control at the Edge and Over 5G},'' in \emph{IEEE International Conference on Edge Computing (EDGE)}, Sept. 2018, pp. 50--57.

\bibitem{E2E_Reliability}
V.~Petrov, M.~A. Lema, M.~Gapeyenko, K.~Antonakoglou, D.~Moltchanov, F.~Sardis, A.~Samuylov, S.~Andreev, Y.~Koucheryavy, and M.~Dohler, ``{Achieving End-to-End Reliability of Mission-Critical Traffic in Softwarized 5G Networks},'' \emph{IEEE Journal on Selected Areas in Communications}, vol.~36, no.~3, pp. 485--501, Mar. 2018.

\bibitem{OROS_WowMom}
C.~Delgado, L.~Zanzi, X.~Li, and X.~Costa-Perez, ``{OROS: Orchestrating ROS-driven Collaborative Connected Robots in Mission-Critical Operations},'' in \emph{IEEE International Symposium on a World of Wireless, Mobile and Multimedia Networks - IEEE WowMom}, June 2022.

\bibitem{OROS_TNSM}
A.~Romero, C.~Delgado, L.~Zanzi, X.~Li, and X.~Costa-Pérez, ``{OROS: Online Operation and Orchestration of Collaborative Robots using 5G},'' \emph{IEEE Transactions on Network and Service Management}, pp. 1--15, 2023.

\bibitem{Singhal2017}
S.~Chetna and S.~D. editors, ``{Resource Allocation in Next-Generation Broadband Wireless Access Networks},'' in \emph{Resource Allocation in Next-Generation Broadband Wireless Access Networks}, 2017.

\bibitem{OSM}
\BIBentryALTinterwordspacing
ETSI, ``{Open source MANO (OSM) project},'' 2024 (Accessed on: June 2024). [Online]. Available: \url{https://www.osm.etsi.org/}
\BIBentrySTDinterwordspacing

\bibitem{ONAP}
\BIBentryALTinterwordspacing
L.~Foundation, ``{Open Network Automation Platform},'' 2024 (Accessed on: June 2024). [Online]. Available: \url{https://www.onap.org/}
\BIBentrySTDinterwordspacing

\bibitem{ROS}
\BIBentryALTinterwordspacing
{Stanford Artificial Intelligence Laboratory et al.}, ``{Robot Operating System},'' 2024 (Accessed on: June 2024). [Online]. Available: \url{https://www.ros.org/}
\BIBentrySTDinterwordspacing

\bibitem{ROS_RCL_APIs}
\BIBentryALTinterwordspacing
{ROS Wiki}, ``{ROS wiki APIs},'' 2024 (Accessed on: June 2024). [Online]. Available: \url{http://wiki.ros.org/APIs/}
\BIBentrySTDinterwordspacing

\bibitem{ROS_RWP_APIs}
\BIBentryALTinterwordspacing
ROS2, ``{ROS Middleware Abstraction Interface },'' 2024 (Accessed on: June 2024). [Online]. Available: \url{https://docs.ros2.org/foxy/api/rmw/}
\BIBentrySTDinterwordspacing

\bibitem{5GT_arch}
{X. Li et al.}, ``{Automating Vertical Services Deployments over the 5GT Platform},'' \emph{IEEE Communications Magazine}, vol.~58, no.~7, pp. 44 -- 50, July 2020.

\bibitem{5Growth-commag}
X.~Li, A.~Garcia-Saavedra, X.~Perez, C.~Bernardos, C.~Guimaraes, K.~Antevski, J.~Mangues, J.~Baranda, E.~Zeydan, D.~Corujo, P.~Iovanna, G.~Landi, J.~Alonso, P.~Paixao, H.~Martins, M.~Lorenzo, J.~Ordo{\~n}ez-Lucena, and D.~L{\'o}pez, ``{5Growth: An End-to-End Service Platform for Automated Deployment and Management of Vertical Services over 5G Networks},'' \emph{IEEE Communications Magazine}, vol.~59, no.~3, pp. 84--90, March 2021.

\bibitem{ifa005}
{ETSI}, ``{ETSI GS NFV-IFA 005 v2.1.1, Network Function Virtualisation (NFV); Management and Orchestration; Or-Vi reference point – Interface and Information Model Specification},'' 2016.

\bibitem{Romero2024}
A.~Romero, C.~Delgado, L.~Zanzi, R.~Suárez, and X.~Costa-Pérez, ``Cellular-enabled collaborative robots planning and operations for search-and-rescue scenarios,'' in \emph{IEEE International Conference on Robotics and Automation (ICRA)}, 2024, pp. 5942--5948.

\bibitem{5TONIC}
5TONIC, ``{5TONIC: Open research and innovation laboratory for 5G Technologies},'' https://www.5tonic.org/, Accessed in June 2024.

\bibitem{bjelonicYolo2018}
M.~Bjelonic, ``{YOLO ROS}: Real-time object detection for {ROS},'' \url{https://github.com/leggedrobotics/darknet_ros}, 2016--2018.

\bibitem{fogros}
J.~Ichnowski, K.~Chen, K.~Dharmarajan, S.~Adebola, M.~Danielczuk, V.~Mayoral-Vilches, H.~Zhan, D.~Xu, R.~Ghassemi, J.~Kubiatowicz \emph{et~al.}, ``{Fogros 2: An adaptive and extensible platform for cloud and fog robotics using ros 2},'' in \emph{Proc. {IEEE} Int. Conf. Robotics and Automation ({ICRA})}.\hskip 1em plus 0.5em minus 0.4em\relax {IEEE}, 2023.

\bibitem{robotkube}
B.~Lampe, L.~Reiher, L.~Zanger, T.~Woopen, R.~van Kempen, and L.~Eckstein, ``{RobotKube: Orchestrating Large-Scale Cooperative Multi-Robot Systems with Kubernetes and ROS},'' in \emph{{2023 IEEE 26th International Conference on Intelligent Transportation Systems (ITSC)}}, 2023, pp. 2719--2725.

\bibitem{kuberos}
Y.~Zhang, C.~Wurll, and B.~Hein, ``{KubeROS: A Unified Platform for Automated and Scalable Deployment of ROS2-based Multi-Robot Applications},'' in \emph{2023 IEEE International Conference on Robotics and Automation (ICRA)}, 2023, pp. 9097--9103.

\bibitem{isaac-sim}
``{NVIDIA Isaac Sim},'' \url{https://developer.nvidia.com/isaac/sim}, accessed: 2024-12-30.

\bibitem{OPENRMF}
``{The Open Robotics Middleware Framework (OPEN-RMF)},'' https://www.open-rmf.org/, Accessed in June 2024.

\bibitem{robomaker}
``{AWS RoboMaker},'' \url{https://aws.amazon.com/es/robomaker/}, accessed: 2024-12-30.

\bibitem{dynorch1}
M.~Schindewolf, D.~Grimm, C.~Lingor, and E.~Sax, ``{Toward a Resilient Automotive Service-Oriented Architecture by using Dynamic Orchestration},'' in \emph{{2022 IEEE 1st International Conference on Cognitive Mobility (CogMob)}}, 2022, pp. 000\,147--000\,154.

\bibitem{dynorch2}
A.~Atutxa, J.~Astorga, M.~Huarte, E.~Jacob, and J.~Unzilla, ``{Enhancing Rescue Operations With Virtualized Mobile Multimedia Services in Scarce Resource Devices},'' \emph{IEEE Access}, vol.~8, pp. 216\,029--216\,042, 2020.

\bibitem{5g-surgery}
\BIBentryALTinterwordspacing
D.~A. Meshram and D.~D. Patil, ``{5G Enabled Tactile Internet for Tele-Robotic Surgery},'' \emph{Procedia Computer Science}, vol. 171, pp. 2618--2625, 2020, third International Conference on Computing and Network Communications (CoCoNet'19). [Online]. Available: \url{https://www.sciencedirect.com/science/article/pii/S1877050920312771}
\BIBentrySTDinterwordspacing

\bibitem{5g-robotics}
\BIBentryALTinterwordspacing
C.~C. Lessi, A.~Gavrielides, V.~Solina, R.~Qiu, L.~Nicoletti, and D.~Li, ``{5G and Beyond 5G Technologies Enabling Industry 5.0: Network Applications for Robotics},'' \emph{{Procedia Computer Science}}, vol. 232, pp. 675--687, 2024, {th International Conference on Industry 4.0 and Smart Manufacturing (ISM 2023)}. [Online]. Available: \url{https://www.sciencedirect.com/science/article/pii/S187705092400067X}
\BIBentrySTDinterwordspacing

\bibitem{robot-dt}
M.~Groshev, C.~Guimarães, A.~De~La~Oliva, and R.~Gazda, ``{Dissecting the Impact of Information and Communication Technologies on Digital Twins as a Service},'' \emph{IEEE Access}, vol.~9, pp. 102\,862--102\,876, 2021.

\bibitem{Voigtlnder2017}
F.~Voigtl{\"a}nder, A.~Ramadan, J.~Eichinger, C.~Lenz, D.~Pensky, and A.~Knoll, ``5{G} for robotics: Ultra-low latency control of distributed robotic systems,'' \emph{International Symposium on Computer Science and Intelligent Controls (ISCSIC)}, pp. 69--72, 2017.

\bibitem{Albonico2021}
M.~Albonico, I.~Malavolta, G.~Pinto, E.~Guzman, K.~Chinnappan, and P.~Lago, ``{Mining Energy-Related Practices in Robotics Software},'' in \emph{Mining Software Repositories Conference (MSR)}, May 2021.

\bibitem{dynamic-task1}
J.~Li, Z.~Cai, M.~Li, W.~Huang, and Y.~Zhang, ``{Dynamic Task Allocation for Heterogeneous Multi-Robot System under Communication Constraints},'' in \emph{{2023 IEEE 6th Information Technology,Networking,Electronic and Automation Control Conference (ITNEC)}}, vol.~6, 2023, pp. 457--463.

\bibitem{dynamic-task2}
\BIBentryALTinterwordspacing
H.~Osooli, P.~Robinette, K.~Jerath, and S.~R. Ahmadzadeh, ``{A Multi-Robot Task Assignment Framework for Search and Rescue with Heterogeneous Teams},'' 2023. [Online]. Available: \url{https://arxiv.org/abs/2309.12589}
\BIBentrySTDinterwordspacing

\bibitem{Rappaport2016}
M.~Rappaport, ``{Energy-aware mobile robot exploration with adaptive decision thresholds},'' \emph{International Symposium on Robotics (ISR)}, pp. 236--243, 2016.

\end{thebibliography}

\vskip -2\baselineskip plus -1fil
\begin{IEEEbiography}[{\includegraphics[width=1in,height=1.25in,clip,keepaspectratio]{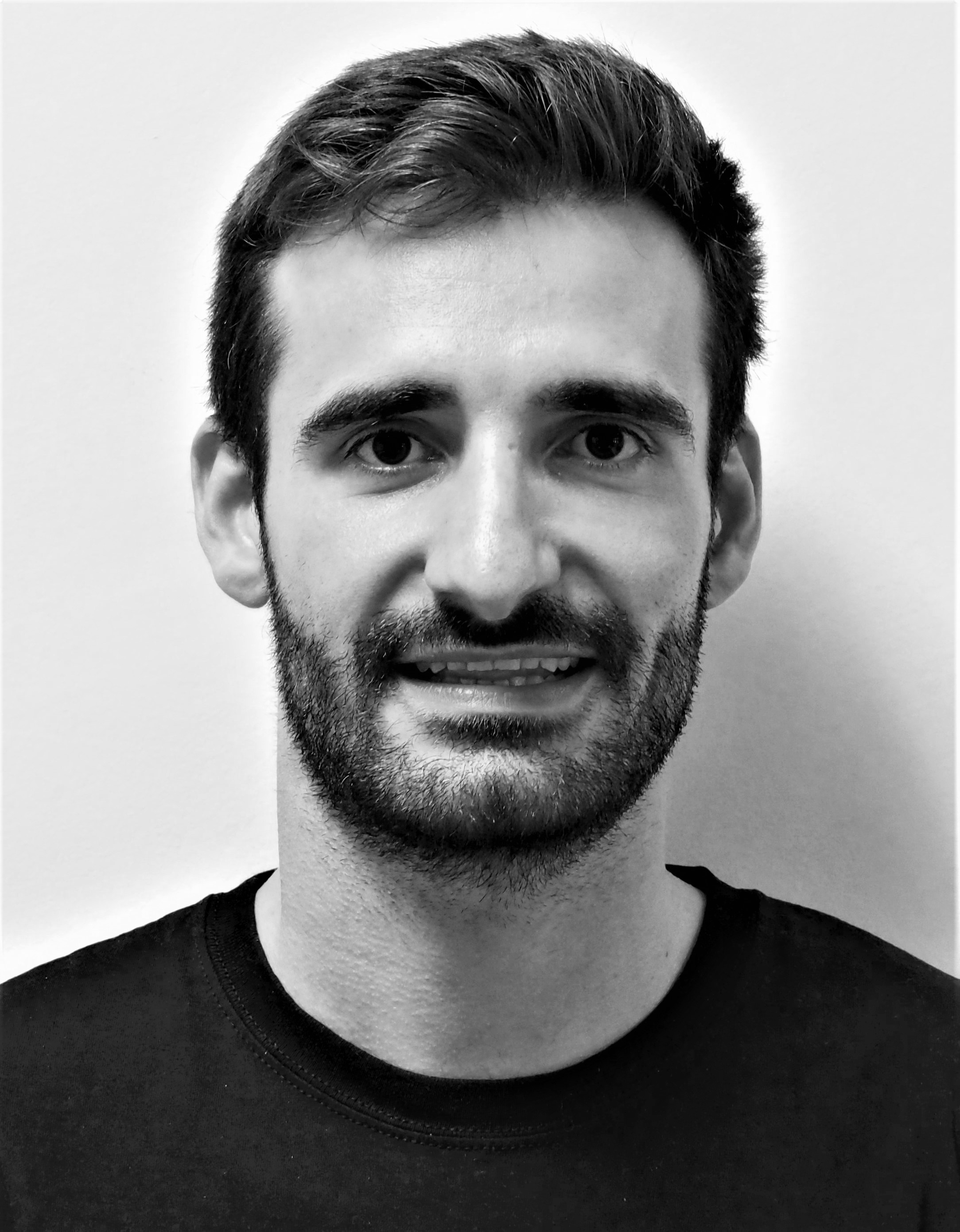}}]{Milan Groshev} received the B.S. degree in telecommunication engineering from the Saints Cyril and Methodius University of Skopje, Macedonia in 2008, the M.S. degree in telecommunication engineering from the Politecnico di Torino, Turin, Italy in 2016 and the  PhD in Telematics Engineering from the University Carlos III Madrid in 2022. He works as a senor researcher at Laude Technology. His research interests include quantum-inspired ML model compression, GNNs in cellular networks and anomaly detection systems.
\end{IEEEbiography}
\vskip -2\baselineskip plus -1fil 
\begin{IEEEbiography}[{\includegraphics[width=1in,height=1.25in,clip,keepaspectratio]{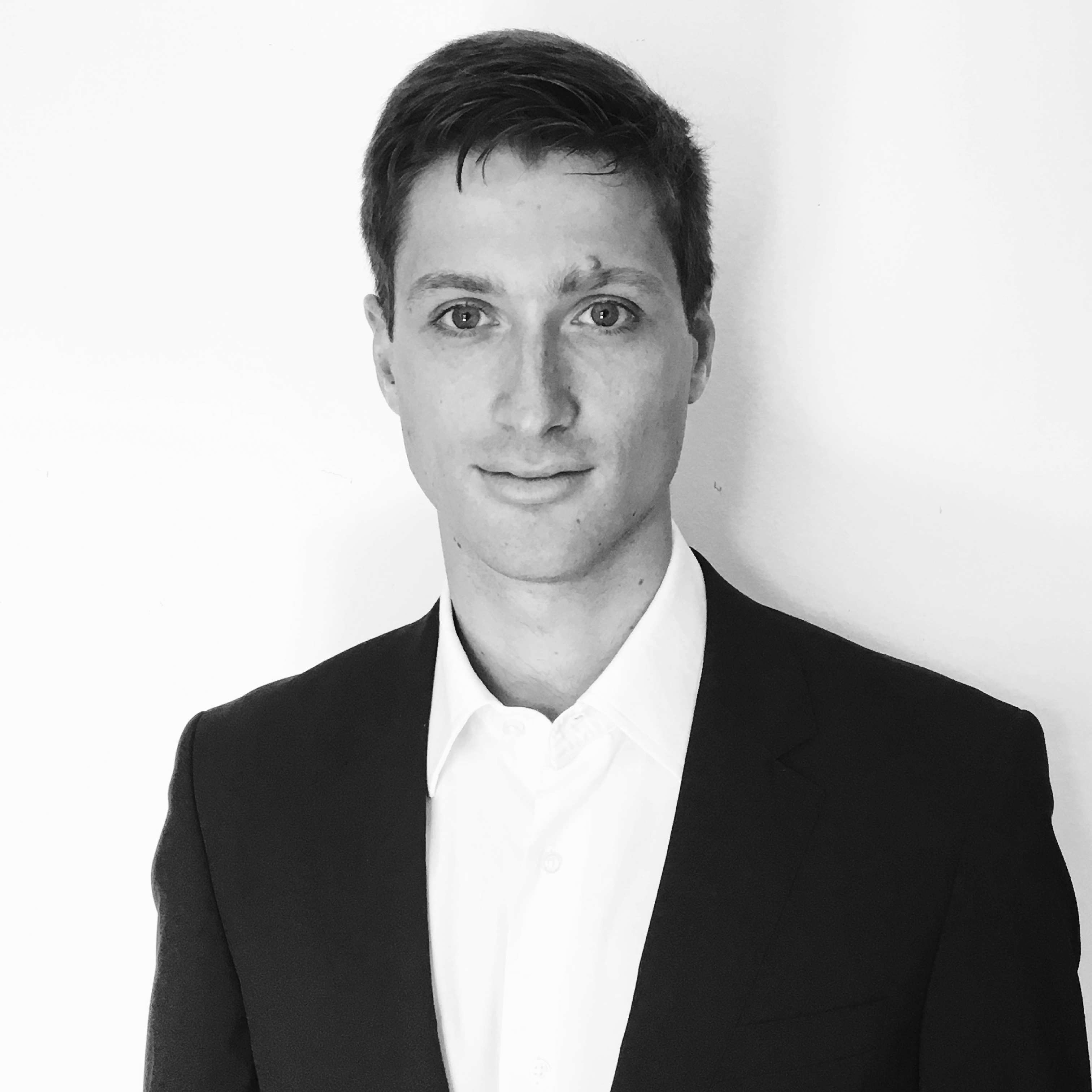}}]{Lanfranco Zanzi} received the B.Sc. and M.Sc. degrees in telecommunication engineering from the Polytechnic of Milan, Italy, in 2014 and 2017, respectively, and the Ph.D. degree from the Technical University of Kaiserlautern, Germany, in 2022. He works as a senior research scientist at NEC Laboratories Europe. His research interests include network virtualization, machine learning, blockchain, and their applicability to 5G and 6G mobile networks.
\end{IEEEbiography}
\vskip -2\baselineskip plus -1fil
\begin{IEEEbiography}[{\includegraphics[width=1in,height=1.25in,clip,keepaspectratio]{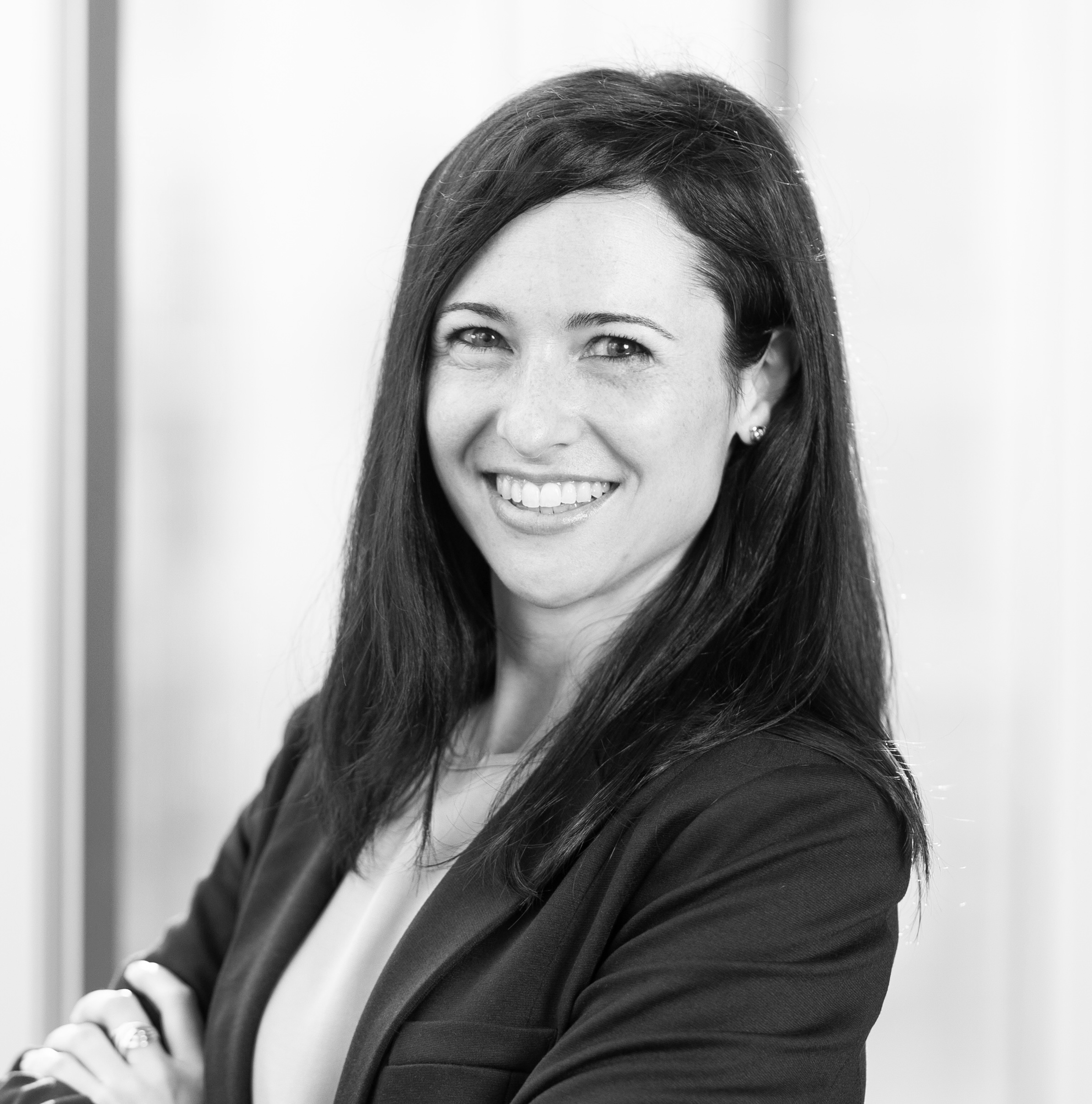}}]{Carmen Delgado} received the M.Sc. in Telecommunications Engineering, M.Sc. in Biomedical Engineering, and the PhD in Telecommunications Engineering from the University of Zaragoza. She works as senior researcher at i2CAT Foundation. 
Her main research interests lie in the field of WSN, IoT, mobile networks, resource allocation, battery-less sensors and communications and Internet of Robotic Things.
\end{IEEEbiography}
\vskip -2\baselineskip plus -1fil
\begin{IEEEbiography}[{\includegraphics[width=1in,height=1.25in,clip,keepaspectratio]{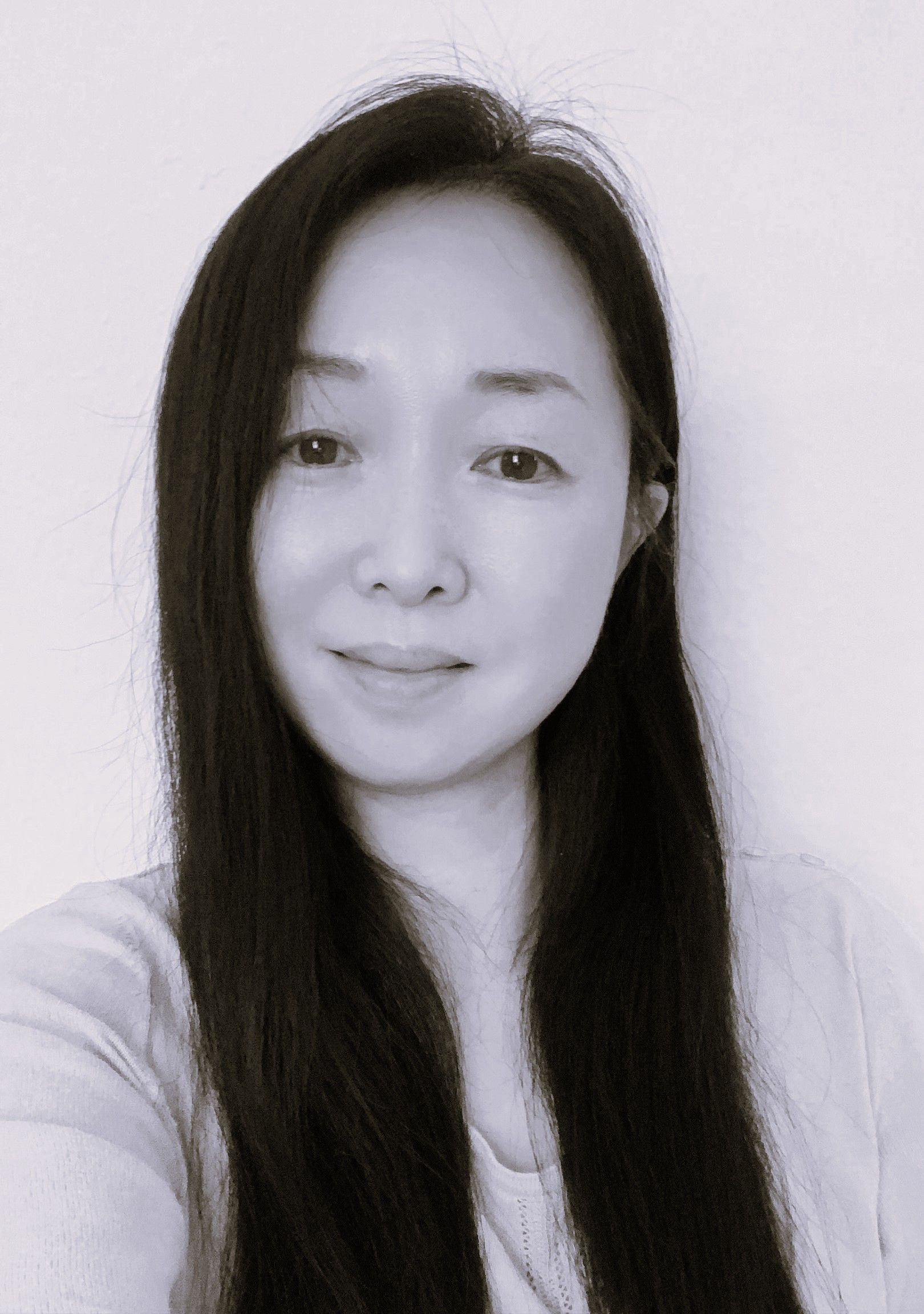}}]{Xi Li} is a Senior Researcher as well as a Program Manager in 6G Networks R\&D at NEC Laboratories Europe, and the Vice Chairman of the SNS-JU 6G Architecture Working Group. She received her M.Sc. in 2002 from the Technical University of Dresden, and Ph.D. in 2009 from University of Bremen, Germany.  Previously, she was a senior researcher fellow and lecturer at the University of Bremen, and a solution designer at Telefonica. She has published high impact publications (Google Scholar Citations 1926 and h-index 25) and owns 11 granted patents, and also an active contributor to IETF CCAMP WG with 3 published RFCs and received best overall award at IETF’99 Hackathon in 2017. 
\end{IEEEbiography}
\vskip -2\baselineskip plus -1fil
\begin{IEEEbiography}[{\includegraphics[width=1in,height=1.25in,clip,keepaspectratio]{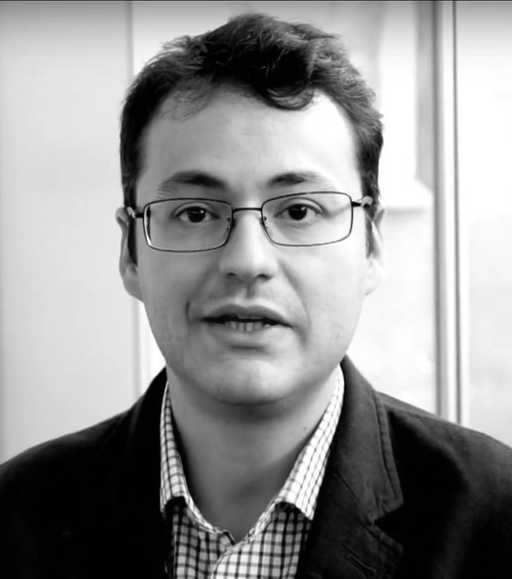}}]{Antonio de la Oliva} Dr. Antonio De La Oliva received his telecommunications engineering degree in 2004 and his Ph.D. in 2008 from the Universidad Carlos III Madrid (UC3M), Spain, where he has been an associate professor since then. 

He is an active contributor to IEEE 802 where he has served as Vice-Chair of IEEE 802.21b and Technical Editor of IEEE 802.21d. He has also served as a Guest Editor of IEEE Communications Magazine. He has published more than 30 papers on different networking areas.
\end{IEEEbiography}
\vskip -2\baselineskip plus -1fil
\begin{IEEEbiography}[{\includegraphics[width=1in,height=1.25in,clip,keepaspectratio]{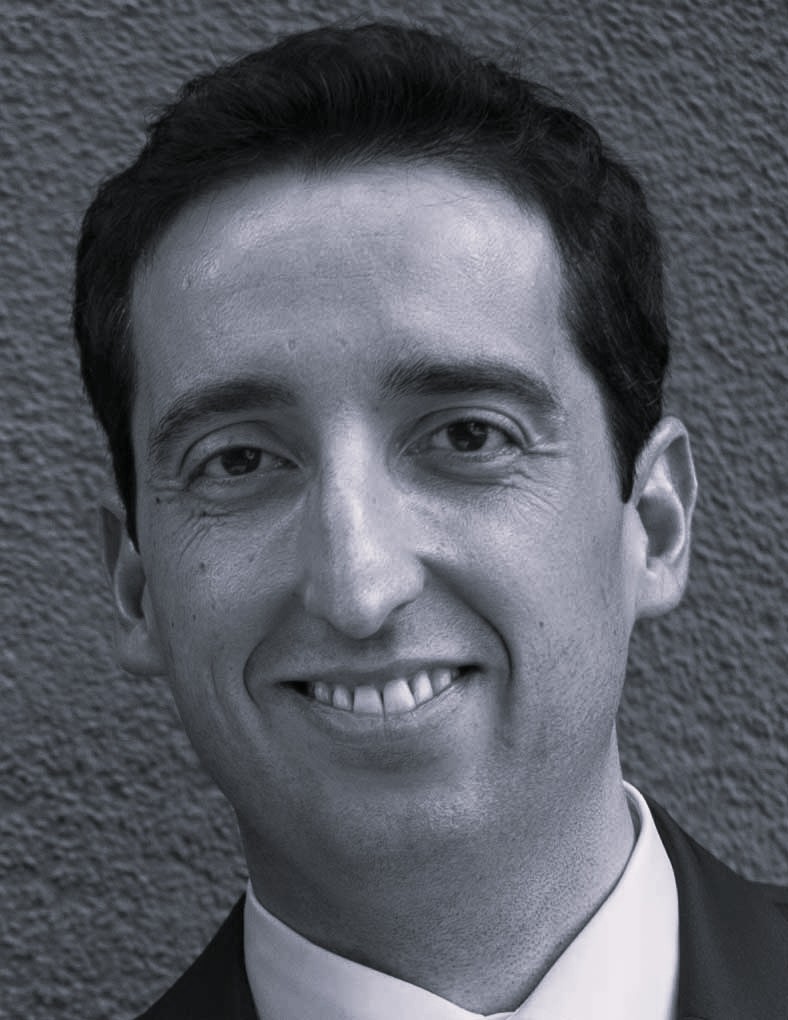}}]
{Xavier~Costa-P\'erez} (M'06--SM'18) is a Research Professor in ICREA, Scientific Director at the i2Cat Research Center and Head of 5G/6G Networks R\&D at NEC Laboratories Europe. He has served on the Organizing Committees of several conferences, published papers of high impact, and holds tenths of granted patents. Xavier received his  Ph.D. degree in Telecommunications from the Polytechnic University of Catalonia (UPC) in Barcelona and was the recipient of a national award for his Ph.D. thesis.
\end{IEEEbiography}

\end{document}